\pdfoutput=1
\documentclass[11pt]{article}

\usepackage[final]{acl}

\usepackage{times}
\usepackage{latexsym}

\usepackage[T1]{fontenc}
\usepackage[utf8]{inputenc}

\usepackage{microtype}

\usepackage{inconsolata}

\usepackage{graphicx}

\usepackage{multirow}
\usepackage{colortbl}  %彩色表格需要加载的宏包
\usepackage{color}
\usepackage{amsmath}
\usepackage{caption}
\usepackage{booktabs}
\usepackage{amssymb}

\usepackage[symbol]{footmisc}
\DefineFNsymbols*{customSymbols}{{\textdagger}{\textasteriskcentered}{\textdaggerdbl}}
\setfnsymbol{customSymbols}

\title{Enhancing Multimodal Unified Representations for Cross Modal Generalization}

\author{
  \textbf{Hai Huang\textsuperscript{1}}\thanks{Equal Contribution},
  \textbf{Yan Xia\textsuperscript{1}}\footnotemark[1],
  \textbf{Shengpeng Ji\textsuperscript{1}},
  \textbf{Shulei Wang\textsuperscript{1}},
  \textbf{Hanting Wang\textsuperscript{1}},
  \textbf{Minghui Fang\textsuperscript{1}}, \\
  \textbf{Jieming Zhu\textsuperscript{2}}\thanks{Project Lead},
  \textbf{Zhenhua Dong\textsuperscript{2}},
  \textbf{Sashuai Zhou\textsuperscript{1}},
  \textbf{Zhou Zhao\textsuperscript{1}}\thanks{Corresponding Author} \\
  \\
  \textsuperscript{1}Zhejiang University, \quad
  \textsuperscript{2}Huawei Noah’s Ark Lab
  \\
  {\texttt{\href{mailto:haihuangcode@outlook.com}{haihuangcode@outlook.com}}}
}

\begin{document}
\maketitle
\begin{abstract}
To enhance the interpretability of multimodal unified representations, many studies have focused on discrete unified representations. These efforts typically start with contrastive learning and gradually extend to the disentanglement of modal information, achieving solid multimodal discrete unified representations. However, existing research often overlooks two critical issues: \textbf{1)} The use of Euclidean distance for quantization in discrete representations often overlooks the important distinctions among different dimensions of features, resulting in redundant representations after quantization; \textbf{2)} Different modalities have unique characteristics, and a uniform alignment approach does not fully exploit these traits. To address these issues, we propose Training-free Optimization of Codebook (TOC) and Fine and Coarse cross-modal Information Disentangling (FCID). These methods refine the unified discrete representations from pretraining and perform fine- and coarse-grained information disentanglement tailored to the specific characteristics of each modality, achieving significant performance improvements over previous state-of-the-art models. The code is available at \href{https://github.com/haihuangcode/CMG}{https://github.com/haihuangcode/CMG}.
\end{abstract}

\section{Introduction}

\label{sec:intro}
Humans' capacity to integrate multimodal information, such as text, audio, and visual, has inspired research on extracting unified information from multimodal data~\citep{harwath2018jointly,miech2019howto100m,shvetsova2022everything,monfort2021spoken}. Researchers aim to develop models that learn unified representations across modalities, using techniques like contrastive learning to map semantically similar multimodal data closer in the embedding space~\citep{radford2021learning,luo2022clip4clip,xu2021videoclip}, achieving notable results in downstream tasks like zero-shot cross-modal retrieval. However, the unbounded nature of the continuous embedding space poses challenges in interpretability. To address this, recent works have explored constructing discrete embedding spaces with prototypes or codebooks, enhancing cross-modal learning and model interpretability~\citep{liu2021cross,lu2022unified,zhao2022towards,xia2024achieving}.

While recent works has demonstrated incredible achievements in multimodal unified representation, there are limitations in terms of the efficiency of embedding space utilization and the granularity of alignment. \textbf{1)} According to previous work~\citep{breiman2001random, wojtas2020feature}, the significance of features varies across different dimensions, and selecting the appropriate dimensions can optimize the feature space, thereby speeding up inference and improving model performance. However, existing multimodal unified representation methods, whether through contrastive learning~\citep{liu2021cross}, teacher-student distillation~\citep{duan2022multi}, or information disentanglement~\citep{xia2024achieving,huang2025semantic}, overlook this issue due to the inherent constraints of the codebook and the quantization method based on Euclidean distance. \textbf{2)} In the unified discrete representation of multimodal data, some studies focus on coarse-grained semantic alignment~\citep{duan2022multi}, others on fine-grained alignment~\citep{xia2024achieving}, and yet others consider both fine and coarse alignments simultaneously~\citep{liu2021cross}. However, these approaches align text with audiovisual data in the same granularity, overlooking the inherent differences between modalities: audiovisual data have temporal fine-grained connections, whereas text represents holistic semantics.

% Previous work, whether through contrastive learning~\citep{liu2021cross}, teacher-student distillation~\citep{duan2022multi}, or information disentanglement~\citep{xia2024achieving}, has aimed to achieve a multimodal discrete unified representation that retains shared information across modalities. However, this shared information still contains redundant background elements that do not contribute to the core semantics. We propose that refining the modal-general features from this perspective could lead to improvements. 

\begin{figure}[t]
  \centering
   \includegraphics[width=1.0\linewidth]{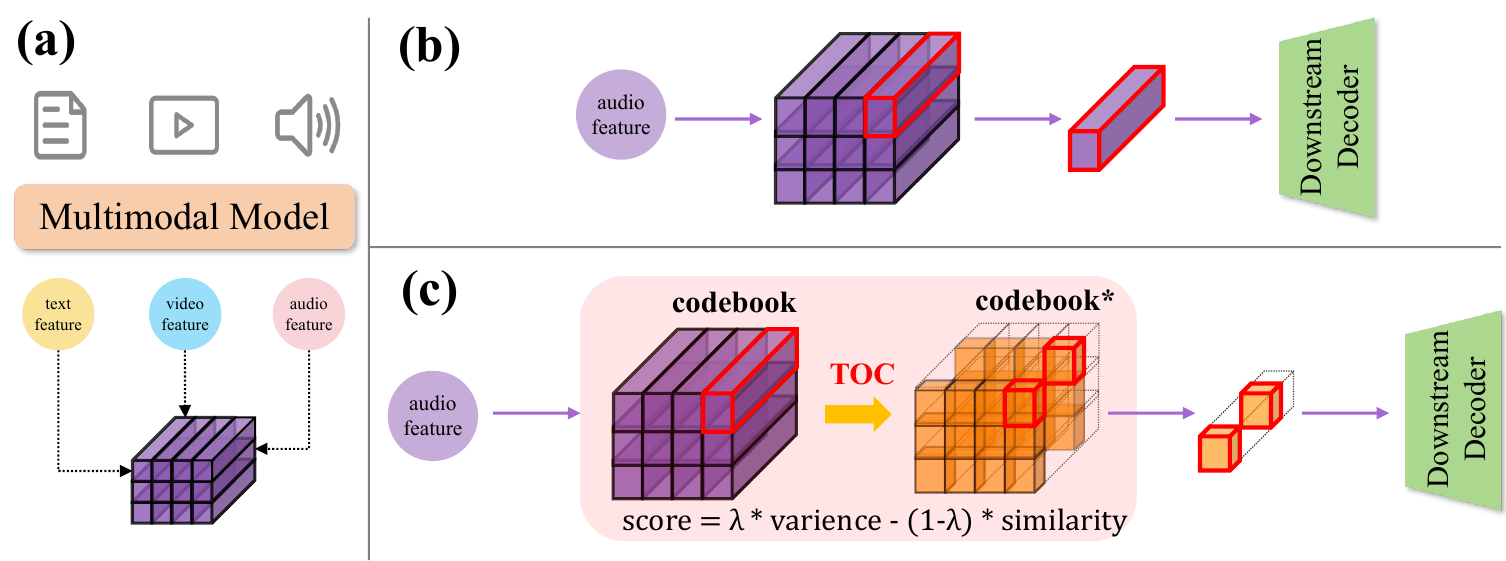}
   \vspace{-5mm}
   \caption{\label{fig:toc}(a) Pretrained multimodal unified discrete representation. 
(b) vanilla downstream experiments using the quantized code from the unified representation. 
(c) After refinement with TOC, downstream experiments are conducted using only a subset of the dimensions. 
%TOC is computed only once and does not require recalculation during subsequent usage, as the codebook remains unchanged after pretraining.
   }
% \vspace{-3mm}
\end{figure}

\begin{figure}[t]
  \centering
   \includegraphics[width=0.95\linewidth]{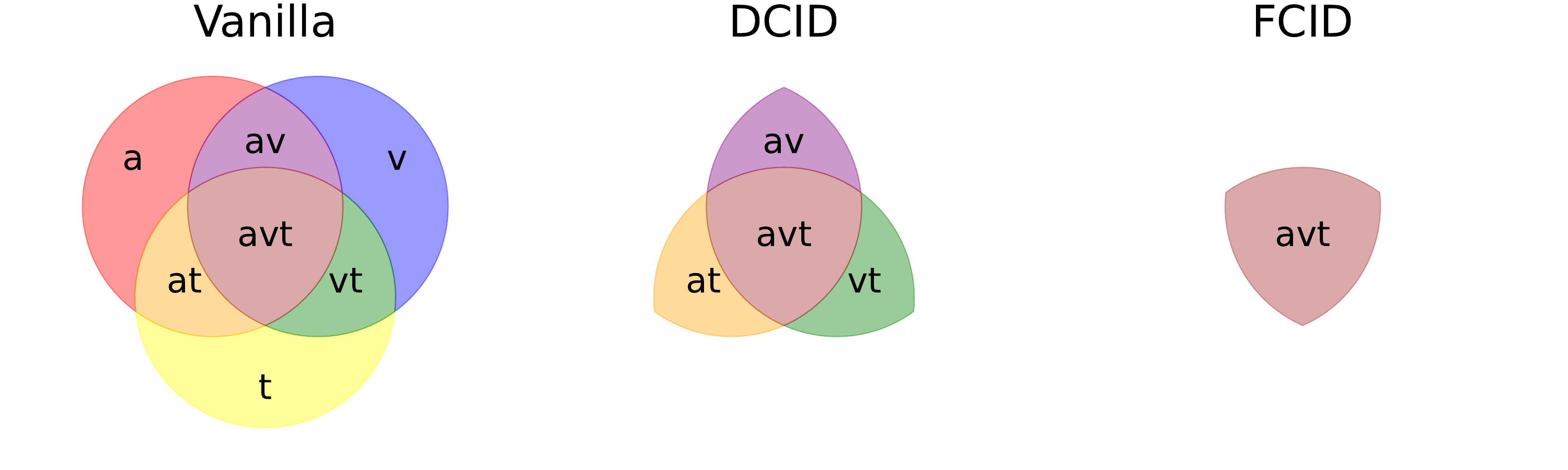}
   \vspace{-2mm}
   \caption{\label{fig:MMUR}Multimodal Unified Representation of Vanilla method, DCID, FCID.
   }
\vspace{-5mm}
\end{figure}

To address the aforementioned issues, we propose two techniques: \textbf{TOC} and \textbf{FCID}. 

\textbf{1)} \textbf{Training-free Optimization of Codebook (TOC)}, inspired by training-free adapters~\citep{zhang2022tip,zhu2023not} and drawing on the concept of feature importance~\citep{breiman2001random, wojtas2020feature, xue2022modality, zhu2023not}, as illustrated in Figure~\ref{fig:toc}, enables the use of a refined code for downstream tasks compared to the conventional approach, which directly uses the full quantized code during inference. The TOC calculation is independent of downstream tasks and does not require additional training. It only needs to be computed once, and the resulting dimensions can be reused for all subsequent inferences.

\textbf{2)} We adjust the order of multimodal alignment based on the inherent differences between text and audio-visual modalities and propose a \textbf{Fine and Coarse Cross-modal Information Disentangling (FCID)} architecture. As shown in Figure~\ref{fig:MMUR}, vanilla methods for learning multimodal unified representations often retain modality-specific information. DCID~\citep{xia2024achieving} introduces disentanglement to separate modality-specific features; however, it does so by repeatedly aligning and decoupling pairs of modalities. For example, when aligning and disentangling audio and text, residual information such as \textit{at} and \textit{avt} is left behind. Similarly, when aligning audio and video, residuals like \textit{av} and \textit{avt} remain. These residuals, \textit{at} and \textit{av}, suggest that the disentanglement process is not fully effective. In contrast, FCID first aligns and disentangles audio and video while retaining fine-grained \textit{av} and \textit{avt} temporal information. It then performs coarse-grained disentangling with text, preserving the shared \textit{avt} information, resulting in a more unified multimodal representation. The main contributions are summarized as follows:
\begin{itemize}
%太长了，具体提升的数据说一下
    \item We propose \textbf{TOC}, a novel method for accurately identifying the importance of feature dimensions through efficient calculations, without the need for additional training. This versatile approach can be seamlessly applied to both multimodal unified and single-modal codebooks, offering a promising and adaptable solution.
    \item We introduce \textbf{FCID}, which disentangles information based on the distinct characteristics of text and audiovisual modalities, preserving both temporal and semantic information across all three modalities, and achieved a more unified multimodal representation.
    \item Our method outperforms the state-of-the-art across various cross-modal generalization tasks. Specifically, FCID, TOC, and their combination improve upon the SOTA by 2.16\%, 1.06\%, and 2.96\%, respectively, across four downstream tasks. Additionally, we validated our approach on zero-shot cross-modal retrieval and cross-modal generation tasks.

    % \item We introduce \textbf{FCID}, which disentangles information based on the distinct characteristics of text and audiovisual modalities, effectively preserving the temporal information of audiovisual data along with the overarching semantic information across all three modalities. 
    % \item Our method significantly outperformed SOTA, across various tasks in the cross-modal generalization setup, showcasing its effectiveness in multimodal learning. Specifically, FCID, TOC, and their combination outperformed before SOTA by 2.16\%, 1.06\%, and 2.96\% respectively, on four downstream tasks. We also validated our method on cross-modal retrieval and generation tasks.
\end{itemize}

\section{Related Work}
\label{sec:appendix_relatedwork}
% In this section, we will introduce recent works on multi-modal unified representations and their distinctions, as well as explorations of training-free methods in other fields. For specific details, please refer to Appendix~\ref{sec:appendix_relatedwork}.

\textbf{Multi-Modal Unified Representation:} Recent work on multi-modal unified representations includes approaches that align modalities into a shared latent space~\citep{petridis2018audio,sarkar2022xkd,andonian2022robust} and train modal-general encoders for cross-modal extraction~\citep{chen2020uniter,wang2022vlmixer}. Cross-modal distillation enables knowledge transfer between modalities~\citep{sarkar2022xkd,pedersoli2022estimating}, while bridging techniques connect representation spaces for improved unified representations~\citep{wang2023connecting}. To enhance interpretability, many works use codebooks or prototypes~\citep{duan2022multi,lu2022unified,liu2021cross,zhao2022towards,xia2024achieving,huang2024unlocking,fang2024ace}. For instance, \citet{duan2022multi} applies Optimal Transport to map features to prototypes, \citet{xia2024achieving} maps multimodal sequences to a common discrete semantic space. Our FCID framework addresses the inherent differences between text and audio-visual modalities through decoupling, enhancing multimodal unified representations.

\noindent
\textbf{Training Free Optimization:} Recent works have explored various training-free methods to boost model performance. Tip-Adapter\citep{zhang2022tip} and APE\citep{zhu2023not} enhance CLIP's few-shot classification, while a training-free method for diffusion models~\citep{chen2024training} optimizes time steps and architecture for efficient image generation. The FuseDream~\citep{liu2021fusedream} combines CLIP and GANs for robust text-to-image generation. TEEN~\citep{wang2024few} offers a training-free solution for few-shot class-incremental learning. SCG-Diffusion~\citep{wang2025towards} proposes a novel training-free method to improve alignment in Transformer-based Text-Guided Diffusion Models. Recent research in video generation~\citep{chen2024delta,yang2024zerosmooth,peng2024conditionvideo,zhang2023controlvideo} and multimodal large language models~\citep{wu2024controlmllm} also focuses on training-free techniques. We introduce TOC, the first training-free optimization method for discrete representation, broadening the scope of training-free approaches in the field.

\section{Backgroud}
\label{sec:appendix_backgroud}
\textbf{Cross Modal Generalization (CMG)} is a task introduced by \citet{xia2024achieving} that evaluates a model's ability to map diverse modalities into a unified discrete latent space. The model's ability for cross-modal zero-shot knowledge transfer is assessed in a setup where training is conducted on modality $m_1$ and testing is performed on modality $m_2$.

During training, the model learns a representation for inputs from one modality using the encoder $\Phi^{m_1}$ and the downstream decoder $\mathbf{D}$:
\begin{equation}
\mathbf{E}(\mathbf{D}(VQ(\Phi^{m_1}(\mathbf{x}^{m_1}_{i}))),\mathbf{y}^{m_1}_{i}),
\end{equation}
where $\mathbf{x}^{m_1}_{i}$ is the input, $\mathbf{y}^{m_1}_{i}$ is the label, and $\mathbf{E}$ is the evaluation function. During testing, the model is evaluated on a different modality $m_2$, demonstrating its ability to generalize:
\begin{equation}
\mathbf{E}(\mathbf{D}(VQ(\Phi^{m_2}(\mathbf{x}^{m_2}_{i}))),\mathbf{y}^{m_2}_{i}).
\end{equation}
Here, $m_1, m_2 \in {a, b, c}$ and $m_1 \neq m_2$. The parameters of both $\Phi^{m_1}$ and $\Phi^{m_2}$ are parameters frozen during training and testing, while only the parameters of $\mathbf{D}$ are updated during training.

\section{Method}
\label{sec:method}
% In this section, we introduce the proposed TOC and FCID, aimed at effectively enhancing the capability of multimodal unified representations. In Section~\ref{subsec:toc}, we introduce the two internal code metrics that constitute TOC. In Sectio~\ref{subsec:FCID}, we elaborate on the foundational principles and design rationale behind the FCID. 

\subsection{Training-free Optimization Codebook}
\label{subsec:toc}
Discrete unified representation spaces commonly employ a codebook structure, where modalities are updated based on the Euclidean distance between their features and the codebook codes. This dimension-equal-weighted update strategy does not consider the varying importance of feature dimensions, leading to redundancy in the final discrete space. According to previous work~\citep{breiman2001random, wojtas2020feature}, the importance of features varies across different dimensions. Eliminating redundant dimensions can help improve performance and accelerate computation.
Therefore, we propose two metrics, Code Similarity and Code Variance, to refine features in the unified space. 

% 与其他training-free方法的区别
%Notably, our approach, TOC, focuses on optimizing the pre-trained codebook and performs calculations independently of downstream information, distinguishing it from methods like Tip-adapter~\citep{zhang2022tip}, APE~\citep{zhu2023not}, them rely on few-shot data to determine the most relevant feature dimensions. Additionally, TOC is designed to tackle more complex downstream tasks, while them are primarily constrained to image classification.

\noindent
\textbf{Code Similarity:} This metric aims to enhance the distinctiveness of codes by extracting feature dimensions that minimize code similarity. We represent the unified representation codebook of modalities as $\mathbf{e} \in \mathbb{R}^{H\times D}$, where $H, D$ denote the size of the codebook and hidden dimension, respectively. 

Assuming the existence of a classification dataset with $C$ categories, acquiring its complete data enables the calculation of the average similarity, denoted as $S$. In an open-world setting, we may assume that the prior probabilities of all categories are equal, denoted as $\frac{1}{C}$. We adopt cosine similarity, $\delta(\cdot, \cdot)$, as the chosen metric:
\begin{equation}
\setlength{\abovedisplayskip}{5pt}
\setlength{\belowdisplayskip}{5pt}
S = \frac{1}{C^{2}} \sum_{i=1}^{C} \sum_{j=1 \atop j \ne i}^{C} \frac{1}{N^{i}N^{j}} \sum_{m=1}^{N^{i}}\sum_{n=1}^{N^{j}} \delta(\mathbf{x}^{i,m},\mathbf{x}^{j,n}),
\end{equation}
where $\mathbf{x}^{i,m}$ and $\mathbf{x}^{j,n}$ denote the input features for the $m$-th and $n$-th samples of categories $i$ and $j$, respectively. $N^i$ and $N^j$ represent their respective total number of training samples.

Each code in the pretrained codebook, $\mathbf{e}^{i} \in \mathbb{R}^{D}, i \in [0, H)$, can be considered as a distinct semantic cluster center, representing a category. Therefore, we can simplify the average similarity calculation:
\begin{equation}
\setlength{\abovedisplayskip}{5pt}
\setlength{\belowdisplayskip}{5pt}
S = \frac{1}{H^{2}} \sum_{i=1}^{H} \sum_{j=1 \atop j \ne i}^{H} \delta(\mathbf{e}^{i},\mathbf{e}^{j}),
\end{equation}

Our goal is to select $Q$ dimensions out of $D$ to enhance the distinctiveness of the codes. We introduce a binary flag $\mathbf{F}\in\left\{0, 1\right \}^{D}$, where $F_k=1$ ($k=1,...,D$) indicates that the $k^{th}$ dimension $\mathbf{e}^{i}_{k}$ is selected, and $\mathbf{F}\mathbf{F}^{\top}=Q$. Our objective now becomes finding the optimal $\mathbf{F}$ to minimize the Code Similarity:

\begin{equation}
% \vspace{3mm}
\setlength{\abovedisplayskip}{5pt}
\setlength{\belowdisplayskip}{5pt}
\min_{\mathbf{F}} \quad S = \frac{1}{H^{2}} \sum_{i=1}^{H} \sum_{j=1 \atop j \ne i}^{H} \delta(\mathbf{e}^{i}\odot \mathbf{F}, \mathbf{e}^{j}\odot \mathbf{F}),
% \vspace{-1mm}
\end{equation}
where $\odot$ denotes element-wise multiplication.

We further suppose the Codebook has been L2-normalized, meaning that each code vector $\mathbf{e}^{i} \in \mathbb{R}^{D}$ has a unit length. Under this assumption, the cosine similarity between two code vectors $\mathbf{e}^{i}$ and $\mathbf{e}^{j}$ can be simplified as their dot product:
\begin{equation}
\delta(\mathbf{e}^{i}, \mathbf{e}^{j}) = \mathbf{e}^{i} \cdot \mathbf{e}^{j},
\end{equation}
where $\cdot$ denotes the dot product of two vectors. Then we can simplify the cosine similarity as
\begin{equation}
\setlength{\abovedisplayskip}{5pt}
\setlength{\belowdisplayskip}{5pt}
S = \sum_{k=d_1}^{d_Q}S_k = \sum_{k=d_1}^{d_Q} \left (  \frac{1}{H^{2}} \sum_{i=1}^{H} \sum_{j=1 \atop j \ne i}^{H} \mathbf{e}^{i}_{k} \cdot \mathbf{e}^{j}_{k} \right ),
\label{equ:optimization_simplify}
\end{equation}
where $k = \left \{d_1, d_2,..., d_Q \right \}$ denotes the indices of selected feature dimensions with ${F}_{k}=1$, and $S_k = \frac{1}{L^{2}} \sum_{i=1}^{L} \sum_{j=1 \atop j \ne i}^{L} \mathbf{e}^{i}_{k} \cdot \mathbf{e}^{j}_{k}$ represents the average inter-class similarity of the $k^{th}$ dimension. Through straightforward derivation, we observe that solving the optimization problem is equivalent to selecting Q elements with the smallest average similarity. 

%---------------------------------
\noindent
\textbf{Code Variance:} Our goal is to reduce redundancy by removing feature dimensions with low variance across codewords, as these dimensions offer minimal discriminative value. The variance for the $k^{th}$ feature dimension is formulated as:
\begin{equation}
\setlength{\abovedisplayskip}{5pt}
\setlength{\belowdisplayskip}{5pt}
V_{k} = \frac{1}{L} \sum_{i=1}^{L} (\mathbf{e}^{i}_{k} - \bar{\mathbf{e}}_{k})^2,
\end{equation}
where $\bar{\mathbf{e}}_{k} = \frac{1}{L} \sum_{i=1}^{L} \mathbf{e}^{i}_{k}$ represents the mean of the $k^{th}$ dimension across all codewords. Similar to Code Similarity, we select the top $Q$ dimensions with the highest variance to enhance discriminative power.

To combine the criteria of similarity and variance, a balance factor $\lambda$ is introduced to compute the final metric for each feature dimension:
\begin{equation}
\label{equ:final_J}
\setlength{\abovedisplayskip}{5pt}
\setlength{\belowdisplayskip}{5pt}
U_k = \lambda V_k - (1-\lambda) S_k,
\end{equation}
where $k = 1, \ldots, D$. The dimensions corresponding to the top-$Q$ biggest values of $U_k$ are chosen as the refined features.
%(-0.7) * sim + 0.3 * torch.var(feats, dim=0)

\begin{figure*}[h]
  \centering
   \includegraphics[width=1.0\linewidth]{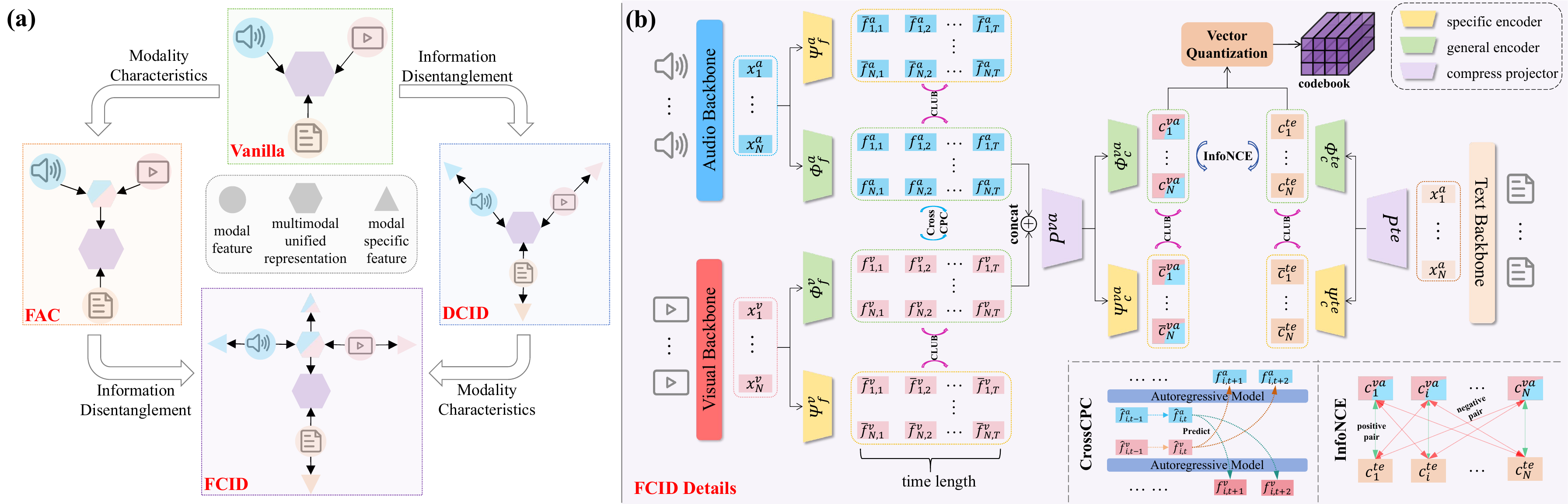}
   % \vspace{-3mm}
   \caption{\label{fig:framwork}
   (a) A simple demonstration of four models: Vanilla method, FAC, DCID, and our proposed FCID. (b) Details of the FCID encoder. 
    On the left, audio and video undergo fine-grained mutual information separation and alignment using modal-general encoders $\Phi^{a}_f, \Phi^{v}_f$ and modal-specific encoders $\Psi^{a}_f, \Psi^{v}_f$. 
    The CLUB module separates specific information $\overline{\mathbf{f}}^{a}_{i}, \overline{\mathbf{f}}^{v}_{i}$ from general information $\mathbf{f}^{a}_{i}, \mathbf{f}^{v}_{i}$, while CrossCPC aligns the general information across modalities. 
    This is followed by compressing the features into unified audiovisual representations. 
    On the right, coarse-grained mutual information separation and alignment are conducted with audiovisual data and text, resulting in a unified discrete representation across all three modalities. 
    The decoder combines the quantized code with modality-specific features and computes the reconstruction loss, omitted in the figure.
   % (a) Vanilla method (b) FAC (c) DCID (d) 我们提出的FCID (e)FCID编码器的细节。On the left, audio and video undergo fine-grained mutual information separation and alignment using modal-general encoders $\Phi^{a}_f, \Phi^{v}_f$ and modal-specific encoders $\Psi^{a}_f, \Psi^{v}_f$. The CLUB module separates specific information $\overline{\mathbf{f}}^{a}_{i}, \overline{\mathbf{f}}^{v}_{i}$ from general information $\mathbf{f}^{a}_{i}, \mathbf{f}^{v}_{i}$, while CrossCPC aligns the general information across modalities. This is followed by compressing the features into unified audiovisual representations. On the right, coarse-grained mutual information separation and alignment are conducted with audiovisual data and text, resulting in a unified discrete representation across all three modalities. 解码器部分负责将量化后的code与modal-specific feature结合在一起与与modal feature计算重构loss，图中省略了这部分。
   }
\vspace{-5mm}
\end{figure*}

%-----------------------------
\subsection{Fine and Coarse cross-modal Information Disentangling}
\label{subsec:FCID}
As shown in Figure~\ref{fig:framwork}, the Vanilla method uses contrastive learning~\citep{liu2021cross} or distillation~\citep{duan2022multi} for cross-modal representation unification. DCID~\citep{xia2024achieving} introduces decoupling to separate modality-specific information. In FAC~\citep{alayrac2020self}, audio and video are first aligned temporally and then with text for semantic alignment. We propose FCID, which combines FAC and DCID. It decouples and aligns audio and video, isolates modality-specific data, and then further aligns with text, discarding non-shared information for a unified multimodal representation. FCID is the first method to simultaneously tackle both modality differences and redundancy issues, inspired by FAC and DCID.

\subsubsection{Fine cross-modal Information Disentangling}
% Given paired audio-video modalities, ${(\mathbf{x}^{a}_{i},\mathbf{x}^{v}_{i})}^{N}_{i=1}$, we employ 
% two fine modal-general encoders $\Phi^{a}_f,\Phi^{v}_f$ to extract fine modal-general features $\mathbf{f}^{a}_{i}, \mathbf{f}^{v}_{i} \in \mathbb{R}^{T \times D}$, and two fine modal-specific encoders $\Psi^{a}_f,\Psi^{v}_f$ to extract fine modal-specific features $\overline{\mathbf{f}}^{a}_{i},\overline{\mathbf{f}}^{v}_{i} \in \mathbb{R}^{T \times D}$, from modalities audio and video, respectively, where $N, T, D$分别代表样本数，音视频序列长度，特征维度:
% \begin{equation}
%     \mathbf{f}^{m}_{i} = \Phi^{m}(\mathbf{x}^{m}_{i}), \ \overline{\mathbf{f}}^{m}_{i} = \Psi^{m}(\mathbf{x}^{m}_{i}), \ m \in \{a,v\}.
% \label{equ1} 
% \end{equation}
Given paired audio-video modalities, ${(\mathbf{x}^{a}_{i},\mathbf{x}^{v}_{i})}^{N}_{i=1}$, we utilize two fine modal-general encoders, $\Phi^{a}_f$ and $\Phi^{v}_f$, to extract fine modal-general features $\mathbf{f}^{a}_{i}$ and $\mathbf{f}^{v}_{i} \in \mathbb{R}^{T \times D}$, and employ two fine modal-specific encoders, $\Psi^{a}_f$ and $\Psi^{v}_f$, to obtain fine modal-specific features $\overline{\mathbf{f}}^{a}_{i}$ and $\overline{\mathbf{f}}^{v}_{i} \in \mathbb{R}^{T \times D}$ from the audio and video modalities, respectively. Here, $N$, $T$, and $D$ represent the number of samples, the length of audio-video sequences, and the feature dimension, respectively. In subsequent equations, $\ m,n \in \{audio,video\}$:

\begin{equation}
    \mathbf{f}^{m}_{i} = \Phi^{m}(\mathbf{x}^{m}_{i}), \ \overline{\mathbf{f}}^{m}_{i} = \Psi^{m}(\mathbf{x}^{m}_{i}).
\label{equ1} 
\end{equation}

Subsequently, we minimize the mutual information between the fine modal-specific features $\mathbf{f}^{m}_{i}$ and $\overline{\mathbf{f}}^{m}_{i}$. At the same time, maximize the mutual information between $\mathbf{f}^{m}_{i}$ and $\mathbf{f}^{n}_{i}$. The details of this approach are outlined below:

\noindent
{\textbf{Mutual Information Minimization: }}CLUB~\citep{cheng2020club} could optimize the mutual information upper bound, demonstrating superior advantages in information disentanglement. Given two variables $\mathbf{x}$ and $\mathbf{y}$, the objective function of CLUB is defined as:
\begin{equation}
\begin{split}
    I_{vCLUB}(\mathbf{x};\mathbf{y}):=\mathbb{E}_{p(\mathbf{x},\mathbf{y}}[\log q_{\theta}(\mathbf{y}|\mathbf{x})] 
    \\-\mathbb{E}_{p(\mathbf{x})}\mathbb{E}_{p(\mathbf{y})}[\log q_{\theta}(\mathbf{y}|\mathbf{x})].
\label{equ4} 
\end{split}
\end{equation}

We use CLUB to optimize the mutual information upper bound between fine modal-general features $\mathbf{f}^{m}_{i}$ and fine modal-specific features $\overline{\mathbf{f}}^{m}_{i}$, where $q_{\theta}$ is the variational approximation of ground-truth posterior of $\mathbf{y}$ given $\mathbf{x}$ and can be parameterized by a network $\theta$. 

% \vspace{-5mm}
\begin{equation}
\begin{split}
    \hat{I}_{vCLUB_f}=\frac{1}{N}\sum_{i=1}^N[\frac{1}{T}\sum_{t=1}^T\log q_{\theta}(\overline{\mathbf{f}}^{m}_{i}|\mathbf{f}^{m}_{i})
    \\- \frac{1}{N}\frac{1}{T}\sum_{j=1}^N\sum_{t=1}^T\log q_{\theta}(\overline{\mathbf{f}}^{m}_{j}|\mathbf{f}^{m}_{i})].
\label{equ5} 
\end{split}
\end{equation}

\noindent
{\textbf{Mutual Information Maximization: }}Contrastive Predictive Coding (CPC)~\citep{oord2018representation} aims to maximize the mutual information between sequence items by predicting future samples using autoregressive models and is widely adopted in self-supervised learning. Given fine general features $\mathbf{f}^{a}, \mathbf{f}^{v} \in \mathbb{R}^{T\times D}$, a prediction horizon of R steps, and a random time moment $t \in (0, \text{T-R}]$, two single-layer unidirectional LSTMs are used to summarize the information of all $\mathbf{f}^{a}_{\leq t}, \mathbf{f}^{v}_{\leq t}$, yielding three context representations as $\mathbf{o}^{m}_{t}$ = LSTM($\mathbf{f}^{m}_{\leq t}$).

%LSTM($\mathbf{f}^{m}_{\leq t} \in \mathbb{R}^{D}, m \in {a,v}$)

For modality M, we first select a set $Z_{neg}$ of N-1 random negative samples and one positive sample $\mathbf{f}^{n}_{t+r}$ from modality N, then use $\mathbf{o}^{m}_{t}$ to predict r-th future step $\mathbf{f}^{n}_{t+r}$ in modality N, and the loss for all modality can be optimized as:

\begin{equation}
\begin{split}
    L^{m2n}_{cpc} = -\frac{1}{R}\sum_{r=1}^R\log\left[\frac{\exp{(\mathbf{f}^{n}_{t+r}W^{m}_{r}\mathbf{o}^{m}_{t})}}{\sum_{\mathbf{f}_{j}\in Z_{neg}}\exp{(\mathbf{f}^{n}_{j}W^{m}_{r}\mathbf{o}^{m}_{t})}}\right].
\label{equ6} 
\end{split}
\end{equation}

\subsubsection{Coarse Cross-modal Information Disentangling}
CCID initially sets up two projectors, $P^{te}$ for compressing textual features and $P^{va}$ for compressing the audiovisual modal-general features obtained from FCID. Subsequently, it configures two coarse modal-specific encoders, $\Psi^{av}_c$ and $\Psi^{te}_c $, to extract coarse modal-specific features $\overline{\mathbf{c}}^{av}_{i}$ and $\overline{\mathbf{c}}^{te}_{i} \in \mathbb{R}^{D}$, and two coarse modal-general encoders, $\Phi^{av}_c$ and $\Phi^{te}_c$, are employed to derive coarse modal-general features $\mathbf{c}^{av}_{i}$ and $\mathbf{c}^{te}_{i} \in \mathbb{R}^{D}$ from the audiovisual and textual modalities, respectively. In subsequent equations, $\mathcal{M}, \mathcal{N}, 
 \in \{audiovisual,text\}$:
\begin{equation}
\begin{split}
    \mathbf{c}^{\mathcal{M}}_{i} = \Phi^{\mathcal{M}}_c(P^{\mathcal{M}}(\mathbf{x}^{\mathcal{M}}_{i})),  \\\overline{\mathbf{c}}^{\mathcal{M}}_{i} = \Psi^{\mathcal{M}}_c(P^{\mathcal{M}}(\mathbf{x}^{\mathcal{M}}_{i})),
\end{split}
\label{equ7}
\end{equation} 

The subsequent process of information disentanglement and alignment is similar to that of Fine cross-modal Information Disentangling.

\noindent
{\textbf{Mutual Information Minimization: }}
We use CLUB to optimize the mutual information upper bound between coarse modal-general features $\mathbf{c}^{\mathcal{M}}_{i}$ and fine modal-specific features $\overline{\mathbf{c}}^{\mathcal{M}}_{i}$, similar to $\hat{I}_{vCLUB_f}$ in FCID:

\begin{equation}
\begin{split}
    \hat{I}_{vCLUB_c}=\frac{1}{N}\sum_{i=1}^N[\frac{1}{T}\sum_{t=1}^T\log q_{\theta}(\overline{\mathbf{c}}^{\mathcal{M}}_{i}|\mathbf{c}^{\mathcal{M}}_{i})\\- \frac{1}{N}\frac{1}{T}\sum_{j=1}^N\sum_{t=1}^T\log q_{\theta}(\overline{\mathbf{c}}^{\mathcal{M}}_{j}|\mathbf{c}^{\mathcal{M}}_{i})].
\label{equ8} 
\end{split}
\end{equation}

\noindent
{\textbf{Mutual Information Maximization: }}Since the coarse information lacks a sequential structure, we transitioned the contrastive learning approach from CPC to InfoNCE, as described below:
\begin{equation}
L_{nce} = -\frac{1}{N} \sum_{i=1}^N 
\log \left[  \frac{\exp(\text{sim}(c_i^{\mathcal{M}}, c_i^{\mathcal{N}})/\tau)}{\sum_{j=1}^N \exp(\text{sim}(c_i^{\mathcal{M}}, c_j^{\mathcal{N}})/\tau)} 
\right].
\end{equation}

\subsection{Final Loss}
Then, we use the codebook to explicitly represent the unified multimodal representation, the latent codebook $\mathbf{e} \in R^{H\times D}$ is shared across modalities audio, video, and text, where $T, H, D$ represent time, size of the discrete latent space, and hidden dimension, respectively. Apply vector quantized operation to map coarse model-general feature $\mathbf{f}^{av}_{i}, \mathbf{f}^{te}_{i}$ to discrete latent codes, $t \in [0, T)$:
\begin{equation}
\begin{split}
    &\hat{\mathbf{c}}^{\mathcal{M}}_{i,t} = VQ(\Phi^{\mathcal{M}}_{c}(\mathbf{x}^{\mathcal{M}}_{i})) = VQ(\mathbf{c}^{\mathcal{M}}_{i,t}) = e_{l}, \\
     &{\rm where} \ l = argmin_{j}\lvert\lvert\Phi_{c}(x) - e_{j}\rvert\rvert_{2}.
% \label{equ2} 
\end{split}
\end{equation}

Then, 
we combine $\mathbf{\hat{c}}_{i}^{m}$ with $\mathbf{\bar{c}}_{i}^{m}$ together to reconstruct original features:
\begin{equation}
\begin{split}
% \label{equ3}
    \underbrace{\|\mathbf{x}_{i}^{\mathcal{M}} - D(\hat{\mathbf{c}}_{i}^{\mathcal{M}};\bar{\mathbf{c}}_{i}^{\mathcal{M}})\|_2^2}_{\textrm{reconstruction loss}} +\underbrace{\beta \|\phi^{\mathcal{M}}_{k}(\mathbf{x}_{i}^{\mathcal{M}}) - \text{sg}[\mathbf{e}]\|_2^2}_{\textrm{commitment loss}},
\end{split}
\end{equation}

and employ Multimodal Exponential Moving Average (MMEMA) strategy to update codebook. 

%The reconstruction loss ensures that the compressed latent codes $e_{l}$ retain the general information of different modalities. Ideally, $\mathbf{z}_{i}^{a}$, $\mathbf{z}_{i}^{b}$, and $\mathbf{z}_{i}^{c}$, encoded from different modalities with the same semantics, should be mapped to the same discrete latent code. 

% However, in the absence of effective supervision, the presence of a modality gap may lead to $\mathbf{z}_{i}^{a}$, $\mathbf{z}_{i}^{b}$, and $\mathbf{z}_{i}^{c}$ converging to distinct regions of the codebook~\citep{zhao2022towards, liu2021cross}. Consequently, we need to minimize the mutual information between the general result and the specific result, as well as maximize the mutual information among the general results of different modalities.

The overall objective of FCID is a combination of these loss functions across both layers:

\begin{equation}
    L = L_{\text{recon}} + L_{\text{commit}} + L_{\text{contra}} + L_{\text{club}},
    \label{eq19}
\end{equation}
where $L_{\text{recon}}$ is the reconstruction loss that merges the modal-specific and modal-general results for each modality and compares them with the original input using MSE loss, $L_{\text{commit}}$ is the commitment loss that computes the MSE loss between the modal-general results and their quantized codes, $L_{\text{contra}} = L_{\text{cpc}} + L_{\text{nce}}$ is the loss that enhances cross-modal alignment and inference by predicting future samples in one modality using information from another, and $L_{\text{club}} = \hat{I}_{vCLUB_f} + \hat{I}_{vCLUB_c}$ represents the mutual information loss concerning the modal-specific and modal-general results within each modality.

\section{Experiment}
\label{sec:experiment}
\subsection{Datasets and Tasks}

\subsubsection{Pretrain}
{\bf Multimodal Unified Representation:}
The pretraining dataset uses VGGsound-AVEL40K~\citep{chen2020vggsound,zhou2022contrastive} with text from~\citet{xia2024achieving}.
{\bf Single-modal Representation:} We trained a VQVAE~\citep{van2017neural} on the CelebA-HQ 30K~\citep{karras2017progressive} dataset and evaluated TOC’s effect on selecting feature dimensions for reconstruction, assessing its transferability to other domains using the codebook.

\subsubsection{Downstream}
The unified representation pre-trained models will be evaluated on several downstream tasks using different datasets. {\bf Cross-modal event classification on AVE dataset:}~\citep{avel} training on one modality and evaluating on another. {\bf Cross-modal event localization on AVVP dataset:}~\citep{tian2020unified} localizing events in one modality and transferring to the other. {\bf Cross-dataset localization/classification:} training on classification in AVE and evaluating localization in AVVP, transferring across datasets. Cross-modal classification between UCF-101~\citep{soomro2012ucf101} visual clips and VGGSound~\citep{chen2020vggsound} audio clips. The decoder in all of the above experiments consists of a single linear layer. {\bf Cross-modal Zero-shot Retrieval:} We adopt a process similar to the test set~\citep{test1k}, which consists of 500 pairs from MSCOCO~\citep{chen2011collecting}, assessing zero-shot retrieval capability for visual-text alignment. Clotho~\citep{drossos2020clotho} assesses zero-shot retrieval capability for audio-text alignment. Flickr Sound~\citep{senocak2018learning} assesses zero-shot retrieval capability for audio-visual alignment. {\bf Cross-modal Generation:} We use a 2-layer MLP and the IP-Adapter~\citep{ye2023ip} as downstream decoders. The MLP is trainable during training but frozen during testing, while the IP-Adapter remains frozen throughout. By leveraging IP-Adapter’s image-to-image capability, the MLP bridges multimodal unified representation and image generation, enabling audio-to-image and text-to-image generation during testing. The model was fine-tuned on 4,500 FlickrSound~\citep{senocak2018learning} image-audio pairs over 80,000 steps with a batch size of 8, and evaluated on 500 additional pairs. Please refer to Appendix~\ref{sec:appendix_imple} for details on the compared works, evaluation metrics, and hyperparameter settings.
\begin{table}[h]
\centering
\resizebox{0.5\textwidth}{!}{
\begin{tabular}{cccccccccc}
\toprule
\multirow{2}{*}{Method} &
\multicolumn{2}{c}{AVE} &
\multicolumn{2}{c}{AVVP} &
\multicolumn{2}{c}{AVE$\rightarrow$AVVP} &
\multicolumn{2}{c}{UCF(v)$\leftrightarrow$VGG(a)} &
\multirow{2}{*}{Avg.} \\
& V$\rightarrow$A & A$\rightarrow$V & V$\rightarrow$A & A$\rightarrow$V & V$\rightarrow$A & A$\rightarrow$V & V$\rightarrow$A & A$\rightarrow$V & \\
\midrule
CODIS     & 36.8 & 39.7 & 32.7 & 32.6 & 40.8 & 40.6 & 50.8 & 45.2 & 39.90\\ 
TURN      & 37.6 & 39.2 & 32.4 & 32.2 & 40.6 & 41.4 & 50.4 & 46.1 & 39.99\\ 
CMCM      & 46.3 & 45.8 & 36.1 & 35.2 & 47.1 & 48.2 & 51.2 & 48.3 & 44.78\\ 
SimMMDG   & 49.5 & 51.7 & 39.3 & 39.7 & 52.9 & 52.7 & 64.5 & 58.8 & 51.14\\ 
DCID      & 54.1 & \textbf{55.0} & 40.4 & 40.8 & 53.0 & 52.4 & 67.1 & 60.6 & 52.93\\ 
FCID      & \textbf{55.2} & 54.9 & \textbf{42.4} & \textbf{44.5} & \textbf{55.3} & \textbf{57.4} & \textbf{69.4} & \textbf{61.6} & \textbf{55.09}\\ 
\midrule
CODIS+TOC & 37.2 & 41.3 & 33.1 & 33.9 & 41.9 & 42.4 & 51.2 & 47.3 & 41.04\color{green}{(+1.14)}\\ 
TURN+TOC  & 38.3 & 40.5 & 33.2 & 32.9 & 41.5 & 43.3 & 51.5 & 46.8 & 41.00\color{green}{(+1.01)}\\ 
CMCM+TOC  & 46.9 & 47.2 & 37.9 & 36.2 & 49.8 & 50.1 & 52.3 & 49.1 & 46.19\color{green}{(+1.41)}\\ 
DCID+TOC  & 54.5 & \textbf{55.0} & 40.9 & 41.6 & 56.5 & 53.6 & 68.1 & 61.7 & 53.99\color{green}{(+1.06)}\\ 
FCID+TOC  & \textbf{55.9} & \textbf{55.0} & \textbf{43.6} & \textbf{45.1} & \textbf{57.4} & \textbf{58.5} & \textbf{69.6} & \textbf{62.0} & \textbf{55.89}\color{green}{(+0.80)}\\ 
\bottomrule
\end{tabular}
}
\caption{Comparison with SOTA Methods on four audiovisual tasks. (SimMMDG represents recent great work in multimodal domain generalization, is incompatible with TOC as it does not use discrete representations.)}
\label{tab:cross-modal generalization}
\vspace{-2mm}
\end{table}

\begin{table*}[t]
\begin{minipage}[t]{0.455\linewidth}
    \centering
    \resizebox{\linewidth}{!}{%
    \begin{tabular}{lcccc}
    \toprule
    Method & V$\leftrightarrow$T (R@10) & A$\leftrightarrow$T (R@10) & V$\leftrightarrow$A (R@10) & Avg. \\
    \midrule
    CMCM      & 7.20  & 14.87 & 15.60 & 7.11 \\
    DCID      & 8.30  & 16.70 & 17.20 & 8.14 \\
    FCID      & \textbf{9.60}  & \textbf{18.19} & \textbf{17.50} & \textbf{8.89} \\
    \midrule
    CMCM+TOC  & 7.70  & 15.33 & 16.10 & 7.52\textcolor{green}{(+0.41)} \\
    DCID+TOC  & 8.80  & 17.08 & 17.80 & 8.56\textcolor{green}{(+0.42)} \\
    FCID+TOC  & \textbf{10.40} & \textbf{19.04} & \textbf{18.40} & \textbf{9.42}\textcolor{green}{(+0.53)} \\
    \bottomrule
    \end{tabular}}
    \caption{Results on three cross-modal zero-shot retrieval tasks, averaged across two directions, with Avg. as the average of R@1, R@5, and R@10 (details in Table~\ref{tab:Details of zero-shot retrieval}).}
    \label{tab:zero-shot retrieval}
\end{minipage}
\hfill
\begin{minipage}[t]{0.26\linewidth}
    \centering
    \resizebox{\linewidth}{!}{%
    \begin{tabular}{cccc}
    \toprule
    Mask (\%) & R100-avg$\downarrow$ & TOC$\downarrow$ & Count$\uparrow$ \\
    \midrule
    87.5 & 0.0621 & 0.0231 & 100 \\
    75.0 & 0.0477 & 0.0159 & 100 \\
    62.5 & 0.0335 & 0.0109 & 100 \\
    50.0 & 0.0229 & 0.0086 & 100 \\
    37.5 & 0.0141 & 0.0062 & 96 \\
    25.0 & 0.0075 & 0.0039 & 90 \\
    \bottomrule
    \end{tabular}}
    \caption{Reconstruction errors under random vs. TOC masking.}
    \label{tab:vqvae_recon}
\end{minipage}
\hfill
\begin{minipage}[t]{0.245\linewidth}
    \centering
    \resizebox{\linewidth}{!}{%
    \begin{tabular}{lccc}
    \toprule
    Method & I2I$\downarrow$ & A2I$\downarrow$ & T2I$\downarrow$ \\
    \midrule
    CMCM       & 129.56 & 130.93 & 148.93 \\
    DCID       & 121.44 & 123.28 & 141.16 \\
    FCID       & \textbf{116.06} & \textbf{117.26} & \textbf{135.52} \\
    \midrule
    CMCM+TOC   & 124.25 & 125.37 & 144.93 \\
    DCID+TOC   & 118.30 & 119.96 & 135.93 \\
    FCID+TOC   & \textbf{113.95} & \textbf{115.14} & \textbf{130.98} \\
    \bottomrule
    \end{tabular}}
    \caption{Cross-modal image generation results (lower is better).}
    \label{tab:gen_result}
\end{minipage}
\vspace{-2mm}
\end{table*}

\begin{figure*}[h]
\begin{minipage}[t]{.48\linewidth}
  \centering
   \includegraphics[width=0.75\linewidth]{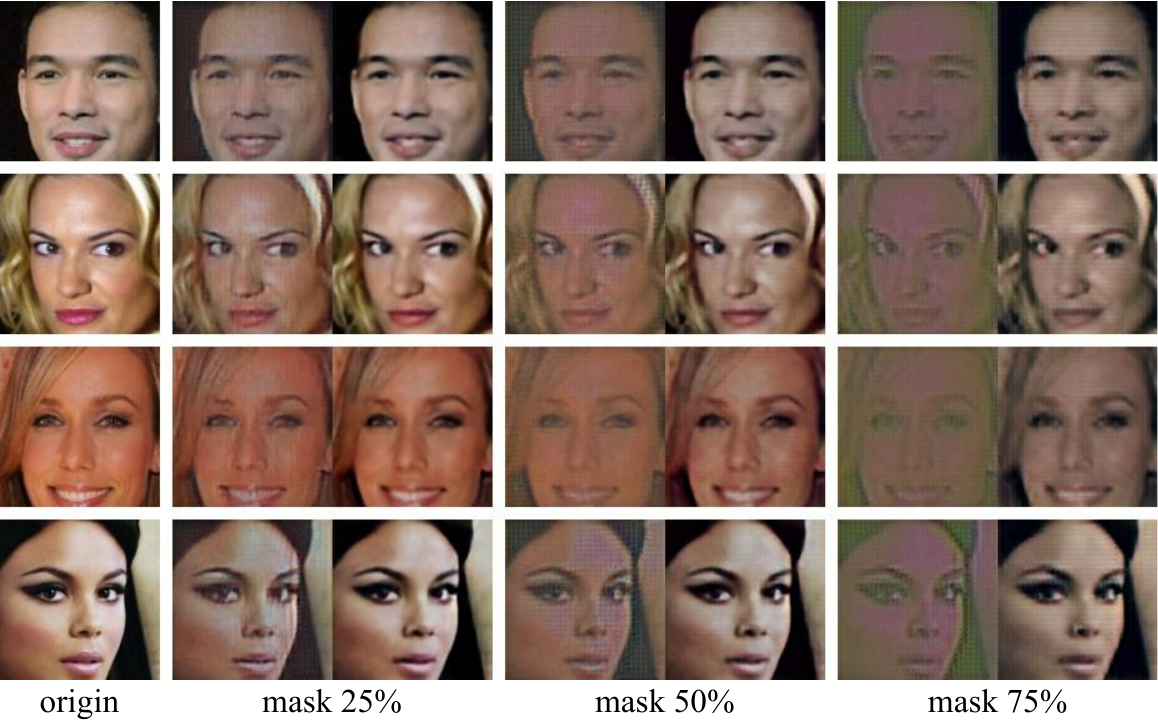}
   \caption{Example results of reconstructions using random and TOC masking.
   \label{fig:vqvae_recon_sample}
   }
\end{minipage}\hfill
\begin{minipage}[t]{.48\linewidth}
  % \centering
  % \resizebox{0.9\linewidth}{!}{
  %   \begin{tabular}{lcc}
  %     \toprule
  %     Method     & I2I↓      & A2I↓      \\ \midrule
  %     CMCM       & 133.56             & 134.59(lower)\\
  %     DCID       & 125.44             & 128.13             \\
  %     FCID     & 123.87             & 125.42             \\ 
  %     CMCM+TOC   & 131.21             & 131.64\\
  %     DCID+TOC   & 123.30              & 123.36             \\ 
  %     FCID+TOC & \textbf{122.29}    & \textbf{122.61}    \\ \bottomrule
  %   \end{tabular}
  % }
  % \captionof{table}{Performance on cross-modal generalization}
  % \label{tab:gen_result}
  \centering
  \includegraphics[width=1\linewidth]{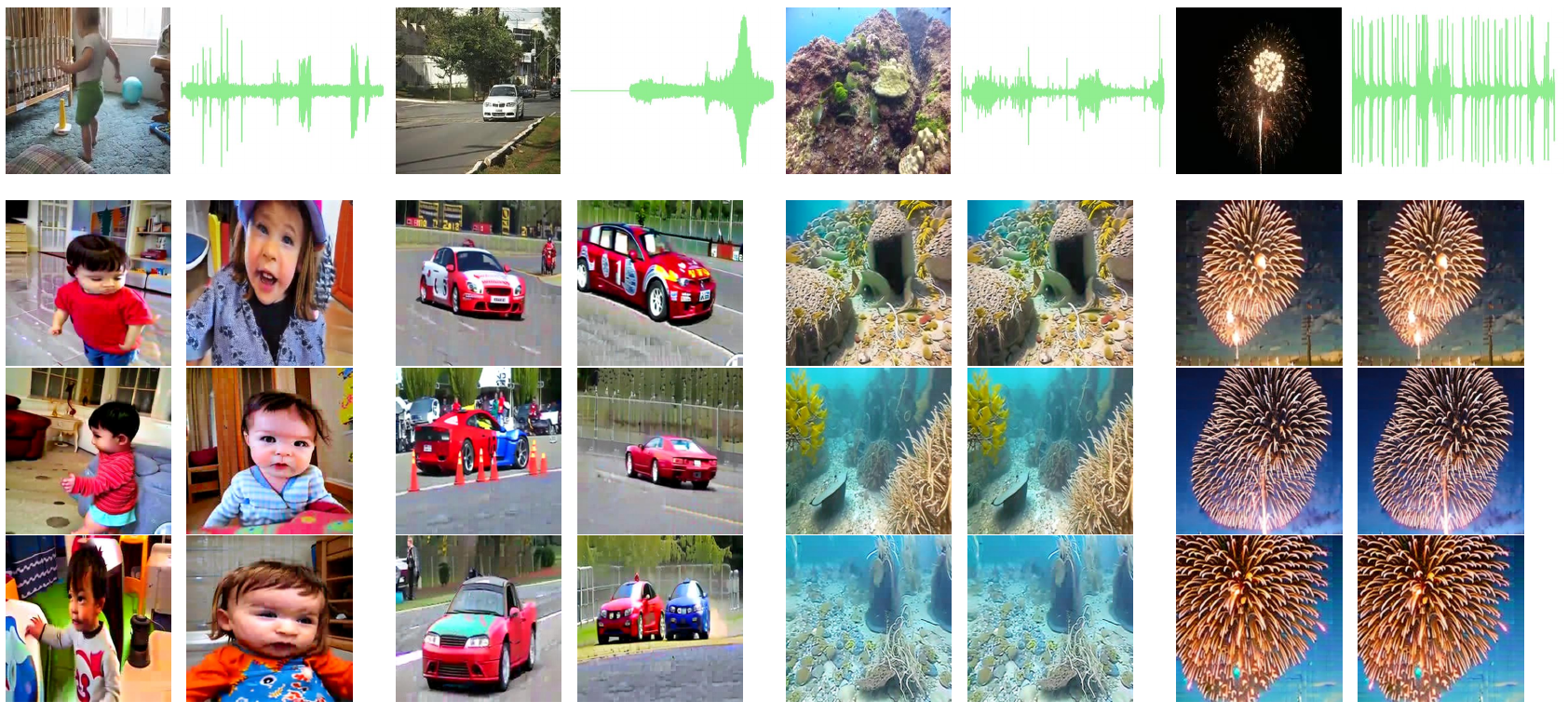}
  \caption{Example results of cross-modal image generation experiments conducted by FCID+TOC.}
  \label{fig:gen_example}
\end{minipage}
\vspace{-5mm}
\end{figure*}

\subsection{Performance Analysis}
\label{sec:preformance_analysis}

In the all tables, \textbf{bold} numbers indicate the best results, while {\color{green}{green}} values in parentheses show the performance improvement attributed to the TOC.

\noindent
\textbf{TOC:} As shown in Table~\ref{tab:cross-modal generalization} and Table~\ref{tab:zero-shot retrieval}, TOC optimizes methods with discrete representation spaces, facilitating at least a 0.80\% improvement in average results for cross-modal generalization tasks, and a minimum average increase of 0.41\% for cross-modal zero-shot retrieval tasks. Additionally, as illustrated in Table~\ref{tab:gen_result} and Figure~\ref{fig:gen_example}, TOC excels in cross-modal generation tasks, improving image-to-image ($I2I$), audio-to-image ($A2I$), and text-to-image ($T2I$) generation performance.

We also explored extending TOC to unimodal discrete representation space. Using a VQVAE model trained on the CelebA-HQ 30K dataset~\citep{karras2017progressive}, we tested reconstructions with a subset of the codeword dimensions. Table~\ref{tab:vqvae_recon} shows that R100-avg represents the average outcome of 100 random codeword dimension selections for reconstruction, where TOC masks the least important dimensions. The MSE for TOC reconstructions with 25\% to 87.5\% of dimensions was significantly lower than the average MSE from random selections. 'Count' represents the number of times the MSE from 100 random selections is greater than the MSE from TOC. Figure~\ref{fig:vqvae_recon_sample} displays reconstruction samples, for all columns except the 'origin' column, with the left half showing reconstructions with randomly masked dimensions and the right half showing TOC-reconstructed images, demonstrating the superior effectiveness of TOC-selected dimensions. For more results, see Appendix~\ref{sec:appendix_recon}.

We also conducted experiments on the impact of TOC on the cosine similarity of the codebook. Please refer to Appendix~\ref{sec:appendix_toc_cosine} for details.

\noindent
\textbf{FCID:} Reviewing Tables~\ref{tab:cross-modal generalization} and~\ref{tab:zero-shot retrieval} clearly shows that FCID and FCID+TOC consistently outperform all other methods across a variety of tasks. Compared to the previous SOTA, FCID achieves an average improvement of 2.16\% in four cross-modal generalization tasks and an average improvement of 0.75\% in three cross-modal zero-shot retrieval tasks. As shown in Table~\ref{tab:gen_result}, these approaches also demonstrate a clear advantage in cross-modal generation tasks. All results suggest that our methods can more effectively process and understand cross-modal information.

As shown in Figure~\ref{fig:gen_example}, the top row displays four image-audio pairs, and the three rows below show images generated from these samples. FCID, fine-tuned only with images, achieves A $\rightarrow$ I results that closely resemble I $\rightarrow$ I outcomes. Notably, in the last two examples, the generated images are identical, indicating that these image-audio pairs map to the same code in the codebook, demonstrating strong modal alignment. For additional T $\rightarrow$ I examples and results, see Appendix~\ref{sec:appendix_gen}.
% As illustrated in Figure~\ref{fig:gen_example}, the top row displays four pairs of image-audio samples, while the three rows below show images generated based on these samples. It is observable that FCID, even when trained only with images, can achieve A $\rightarrow$ I results, closely resembling the I $\rightarrow$ I outcomes, especially in the last two examples where the generated images are identical. This indicates that these two pairs of image-audio samples are mapped to the same code in the codebook, demonstrating a high degree of modal alignment. For additional examples and results for T $\rightarrow$ I, please refer to Appendix~\ref{sec:appendix_gen}.

% We further demonstrate multimodal quantization activations in the discrete representation spaces of DCID~\citep{xia2024achieving} and FCID. Using tri-modal audio-video-text data from VALOR32K~\citep{chen2023valor}, we quantify the code activations in the DCID and FCID codebooks, as shown in Figures~\ref{fig:dcid_codebook} and~\ref{fig:FCID_codebook}. In these figures, {\color{red}{red}} points indicate a single modality activation > 95\%, {\color{green}{green}} points show activation across all three modalities $\geq$ 5\%, and {\color{blue}{blue}} points fall between these categories. The results clearly show that FCID achieves much better alignment across the three modalities than DCID, with fewer codes activated by a single modality. This demonstrates the improvement FCID offers for unified tri-modal representations.

\begin{figure}[t]
\begin{minipage}[t]{.48\linewidth}
  \centering
   \includegraphics[width=1.0\linewidth]{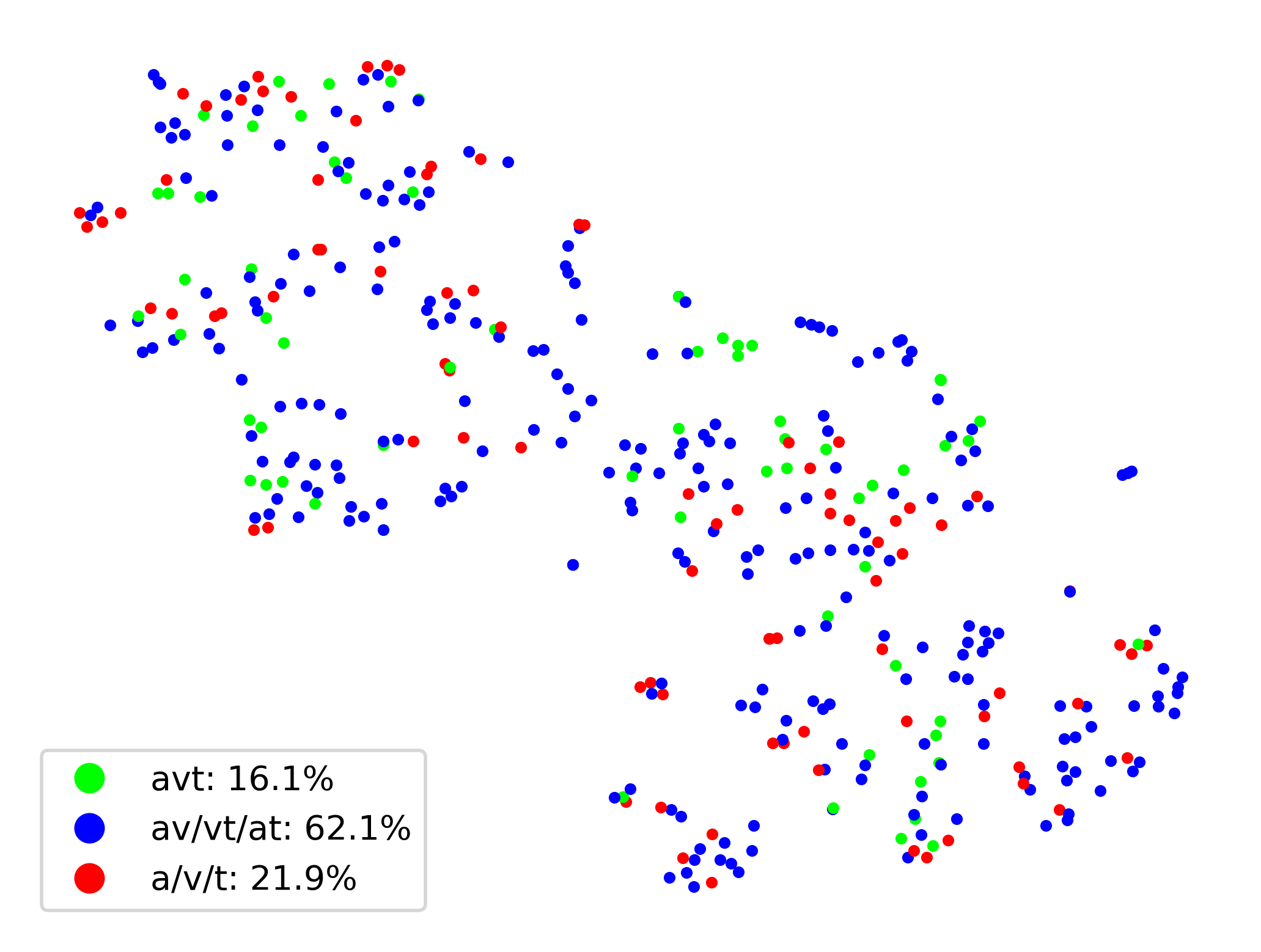}
   \caption{DCID's codebook activate
   \label{fig:dcid_codebook}
   }
\end{minipage}\hfill
\begin{minipage}[t]{.48\linewidth}
  \centering
  \includegraphics[width=1.0\linewidth]{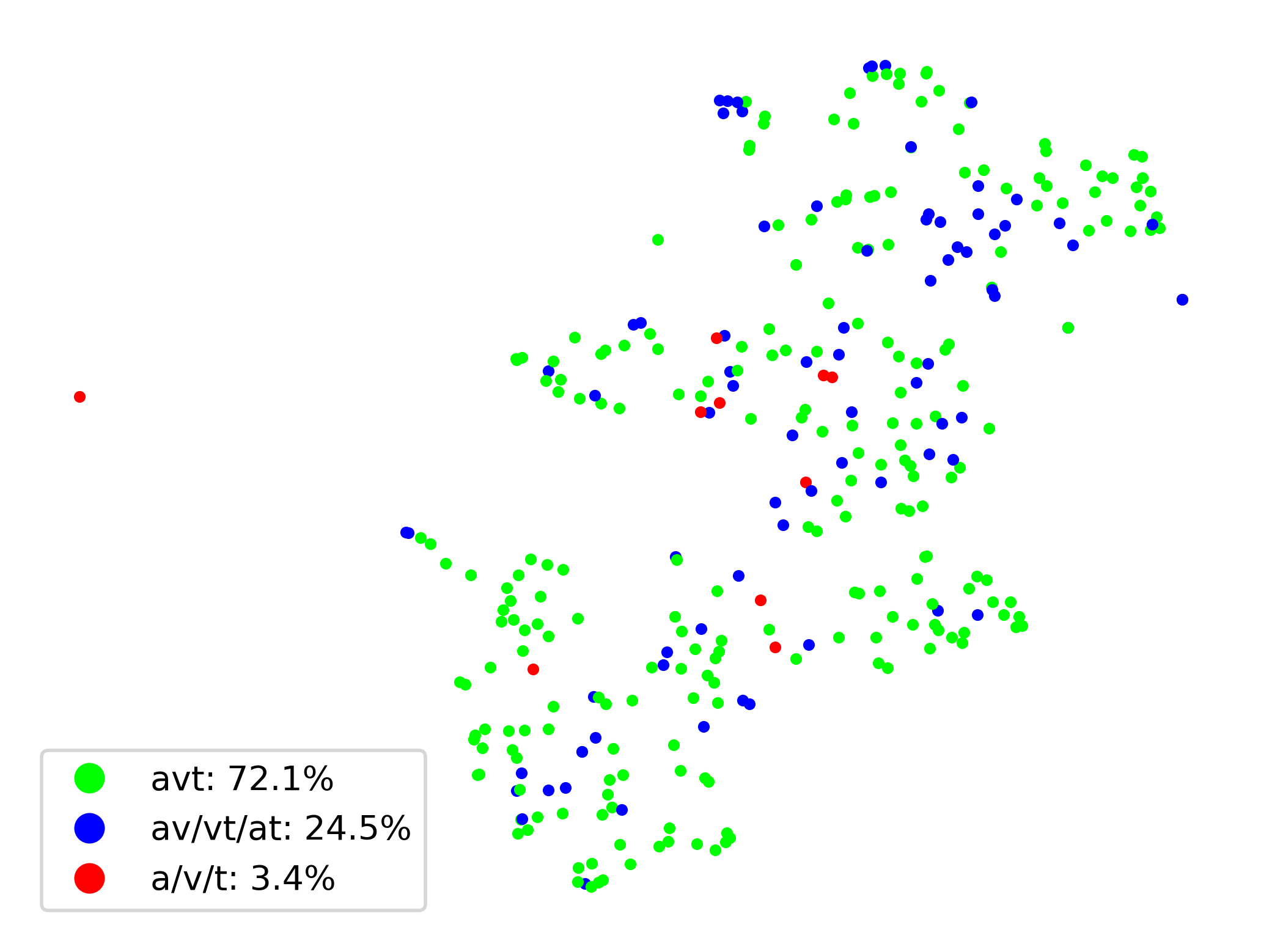}
  \caption{FCID's codebook activate}
  \label{fig:FCID_codebook}
\end{minipage}
\vspace{-5mm}
\end{figure}

We demonstrate multimodal quantization activations in the discrete representation spaces of DCID~\citep{xia2024achieving} and FCID using tri-modal audio-video-text data from VALOR32K~\citep{chen2023valor}. Figures~\ref{fig:dcid_codebook} and~\ref{fig:FCID_codebook} show the code activations in the DCID and FCID codebooks. {\color{red}{Red}} points indicate single modality activations > 95\%, {\color{green}{green}} points show activation across all three modalities $\geq$ 5\%, and {\color{blue}{blue}} points fall between these categories. More {\color{green}{green}} dots indicate closer alignment of discrete representations across modalities in the codebook, while more {\color{red}{red}} dots reflect greater divergence. FCID outperforms DCID in aligning the three modalities, with fewer codes activated by a single modality, demonstrating its improved ability to learn unified tri-modal representations.

For details on the computational efficiency and loss variation of FCID pre-training, please refer to Appendix~\ref{sec:appendix_computational_efficiency} and~\ref{sec:appendix_loss}, respectively.
% For details on loss variation of FCID pre-training, please refer to Appendix~\ref{sec:appendix_loss}.

\subsection{Ablation Study}
\label{sec:ablation_study}
\textbf{Loss: }We conducted ablation experiments on all losses in Equation~\ref{eq19}, excluding the contrastive loss, as it is fundamental to multimodal alignment in this work. Without contrastive learning, alignment cannot be achieved, rendering its ablation unnecessary. As shown in Table~\ref{tab:ablation of losses}, we observe that $L_{commit}$ has a minor impact on results, as its primary role is to align unified discrete representations with quantized features. In contrast, $L_{recon}$ is crucial for ensuring semantic completeness after disentanglement, minimizing information incompleteness during this process. Finally, $L_{club}$ directly evaluates the success of information disentangling, and its absence significantly degrades model performance.

% \begin{table}[h]
% % \vspace{-3mm}
% \centering
% \resizebox{0.4\textwidth}{!}{%
% \begin{tabular}{lccccccccc}
% \toprule
% \centering
% \multirow{1}{*}{Method}                                                                                               \\   
%                                  &
%                                  \multicolumn{2}{c}{\begin{tabular}[c]{@{}c@{}}AVE\\ V$\rightarrow$A   A$\rightarrow$V\end{tabular}} & 
%                                  \multicolumn{2}{c}{\begin{tabular}[c]{@{}c@{}}AVVP\\ V$\rightarrow$A   A$\rightarrow$V\end{tabular}} & 
%                                  \multicolumn{2}{c}{\begin{tabular}[c]{@{}c@{}}AVE$\rightarrow$AVVP\\ V$\rightarrow$A   A$\rightarrow$V\end{tabular}} & 
%                                  \multicolumn{2}{c}{\begin{tabular}[c]{@{}c@{}}UCF(v)$\leftrightarrow$VGG(a)\\ V$\rightarrow$A   A$\rightarrow$V\end{tabular}}&
%                                  \multicolumn{1}{c}{\begin{tabular}[c]{@{}c@{}}Avg.\\ \end{tabular}}\\
%                                  \hline
% Ours   &\textbf{55.9} & \textbf{55.0} & \textbf{43.6} & \textbf{45.1} & \textbf{57.4} & \textbf{58.5} & \textbf{69.6} & 62.0 & \textbf{55.89}\\ 
% w/o $L_{club}$   &51.3 & 51.6 & 39.5 & 40.7 & 50.6 & 51.1 & 63.3 & 57.6 & 50.71\\ 
% w/o $L_{recon}$   &53.3	&53.6	&42.5	&43.3	&56.1	&57.5	&66.8	&\textbf{62.3}	&54.43\\ 
% w/o $L_{commit}$   &54.6	&54.8	&42.9	&44.7	&57.2	&56.8	&67.9	&61.4	&55.04\\ 
% \bottomrule
% \end{tabular}%
% }
% \caption{Ablation studies on the impact of different loss.}
% \label{tab:ablation of losses}
% \vspace{-4mm}
% \end{table}
\begin{table}[h]
\centering
\resizebox{0.48\textwidth}{!}{%
\begin{tabular}{lccccccccc}
\toprule
Method &
\multicolumn{2}{c}{\begin{tabular}[c]{@{}c@{}}AVE\\ V$\rightarrow$A \quad A$\rightarrow$V\end{tabular}} &
\multicolumn{2}{c}{\begin{tabular}[c]{@{}c@{}}AVVP\\ V$\rightarrow$A \quad A$\rightarrow$V\end{tabular}} &
\multicolumn{2}{c}{\begin{tabular}[c]{@{}c@{}}AVE$\rightarrow$AVVP\\ V$\rightarrow$A \quad A$\rightarrow$V\end{tabular}} &
\multicolumn{2}{c}{\begin{tabular}[c]{@{}c@{}}UCF(v)$\leftrightarrow$VGG(a)\\ V$\rightarrow$A \quad A$\rightarrow$V\end{tabular}} &
Avg. \\
\midrule
Ours           & \textbf{55.9} & \textbf{55.0} & \textbf{43.6} & \textbf{45.1} & \textbf{57.4} & \textbf{58.5} & \textbf{69.6} & 62.0 & \textbf{55.89} \\
w/o $L_{club}$   & 51.3 & 51.6 & 39.5 & 40.7 & 50.6 & 51.1 & 63.3 & 57.6 & 50.71 \\
w/o $L_{recon}$  & 53.3 & 53.6 & 42.5 & 43.3 & 56.1 & 57.5 & 66.8 & \textbf{62.3} & 54.43 \\
w/o $L_{commit}$ & 54.6 & 54.8 & 42.9 & 44.7 & 57.2 & 56.8 & 67.9 & 61.4 & 55.04 \\
\bottomrule
\end{tabular}%
}
\caption{Ablation studies on the impact of different loss.}
\label{tab:ablation of losses}
\vspace{-4mm}
\end{table}

\noindent
\textbf{TOC: }As presented in Table~\ref{tab:Ablation of TOC}, the two components of TOC, when applied individually, led to improvements in FCID performance by 0.54\% and 0.10\% across eight metrics, respectively. The combined effect of both components resulted in a total improvement of 0.80\%. Refining the discrete representation space using either Code Similarity or Code Variance alone effectively reduces feature redundancy and enhances model performance in downstream tasks. However, Code Similarity demonstrates a greater impact than Code Variance. When both components are combined, the best performance is achieved, highlighting the effectiveness of our proposed TOC design.

\begin{table}[h]
% \vspace{-3mm}
\centering
\resizebox{0.48\textwidth}{!}{%
\begin{tabular}{cccccccccccc}
\toprule
\centering
\multirow{1}{*}{{S }} & \multirow{1}{*}{{R }} 
                                 & 
                                 \multicolumn{2}{c}{\begin{tabular}[c]{@{}c@{}}AVE\\ V$\rightarrow$A   A$\rightarrow$V\end{tabular}} & 
                                 \multicolumn{2}{c}{\begin{tabular}[c]{@{}c@{}}AVVP\\ V$\rightarrow$A   A$\rightarrow$V\end{tabular}} & 
                                 \multicolumn{2}{c}{\begin{tabular}[c]{@{}c@{}}AVE$\rightarrow$AVVP\\ V$\rightarrow$A   A$\rightarrow$V\end{tabular}} & 
                                 \multicolumn{2}{c}{\begin{tabular}[c]{@{}c@{}}UCF(v)$\leftrightarrow$VGG(a)\\ V$\rightarrow$A   A$\rightarrow$V\end{tabular}}&
                                 \multicolumn{1}{c}{\begin{tabular}[c]{@{}c@{}}Avg.\\ \end{tabular}}\\
                                 \hline
- & - &55.2 & 54.9 & 42.4 & 44.5 & 55.3 & 57.4 & 69.4 & 61.6 & 55.09 \\
\checkmark & - &55.8 & 54.5 & \textbf{43.6} & \textbf{45.7} & 56.8 & 58.3 & 69.2 & 61.1 & 55.63 \\
-&\checkmark & 55.6 & \textbf{55.0} & 43.4 & 44.8 & 56.2 & 54.8 & \textbf{69.8} & 61.9 & 55.19 \\
\checkmark&\checkmark & \textbf{55.9} & \textbf{55.0} & \textbf{43.6} & 45.1 & \textbf{57.4} & \textbf{58.5} & 69.6 & \textbf{62.0} & \textbf{55.89} \\ 
\bottomrule
\end{tabular}%
}
\caption{Ablation studies on the impact of TOC (S and R represent Code Similarity and Code Variance, respectively)}
\label{tab:Ablation of TOC}
\vspace{-3mm}
\end{table}

\noindent
\textbf{FCID: }We focus our ablation studies on the key disentanglement components, $\hat{I}{vCLUB_f}$ and $\hat{I}{vCLUB_c}$, which involve $A_{CLUB}, V_{CLUB}$ and $AV_{CLUB}, TE_{CLUB}$, respectively. Table~\ref{tab:Ablation of FCID} shows that $A_{CLUB}$ and $V_{CLUB}$ significantly impact audiovisual-related downstream tasks. $TE_{CLUB}$ also influences results, as improperly disentangled textual information can affect performance by containing irrelevant or missing AV data. Similarly, using only $AV_{CLUB}$ still retains some modality-specific information, but the disentanglement and alignment with text help separate these features.

\begin{table}[h]
% \vspace{-3mm}

\centering
\resizebox{0.36\textwidth}{!}{%
\begin{tabular}{cccccc}
\toprule
\centering
\multirow{1}{*}{{$A_{CLUB}$}} 
& \multirow{1}{*}{{$V_{CLUB}$}} 
& \multirow{1}{*}{{$AV_{CLUB}$}} 
& \multirow{1}{*}{{$TE_{CLUB}$}} 
&
                                 \multicolumn{1}{c}{\begin{tabular}[c]{@{}c@{}}Avg.\\ \end{tabular}}\\
                                 \hline
- & - &- & - & 50.71\\ 
\checkmark & - &- & - &52.79\\ 
- & \checkmark &- & - &53.40\\ 
- & - &\checkmark & - &51.78\\ 
- & - &- & \checkmark &51.59\\ 
\checkmark & \checkmark &- & -  &54.34\\ 
- & - &\checkmark & \checkmark  &52.46\\ 
\checkmark & \checkmark &\checkmark & \checkmark &\textbf{55.09}\\ 
\bottomrule
\end{tabular}%
}
\caption{Ablation studies on the impact of FCID (detailed results for each task are provided in Table~\ref{tab:Details of Ablation of FCID})}
\label{tab:Ablation of FCID}
\vspace{-5mm}
\end{table}

% \subsection{Ablation Study}

\section{Conclusion}
\label{sec:conclusion}
Inspired by feature importance and training-free optimization, we propose TOC, the first training-free optimization method for discrete representation space, enhancing both multimodal and single-modal representations. We also introduce FCID, a framework that integrates disentanglement with modality-specific characteristics to achieve fine-grained audio-video temporal alignment and coarse-grained text semantic alignment. This disentanglement separates modality-specific information, yielding a unified multimodal representation. Extensive experiments on cross-modal classification, localization, retrieval, and generation tasks validate the effectiveness of our approach.

% Inspired by feature importance and training-free optimization, we propose TOC, the first application of training-free optimization in the discrete representation space, enhancing both multimodal and single-modal representations. Additionally, we introduce the FCID framework, which integrates disentanglement with modality-specific characteristics to enable fine-grained temporal alignment between audio and video and coarse-grained semantic alignment with text. This disentanglement effectively separates modality-specific information, resulting in a unified multimodal representation. Extensive experiments across tasks such as cross-modal classification, localization, retrieval, and generation demonstrate the effectiveness of our approach.

% Inspired by works on feature importance and training-free optimization, we propose TOC. This is the first application of training-free optimization to the discrete representation space, enhancing multimodal and single-modal (e.g., images) representations. We also introduce the FCID framework. By combining disentanglement with the intrinsic characteristics of each modality, FCID incorporates fine-grained temporal alignment between audio and video, as well as coarse-grained semantic alignment between audiovisual data and text. Disentanglement is applied throughout this process to separate modality-specific information, achieving a more unified multimodal representation. Experiments conducted on tasks such as cross-modal classification, localization, retrieval, and generation demonstrate the effectiveness of our proposed method.

\section*{Limitations}
\label{sec:limitations}
\textbf{TOC: }We assumed equal prior probabilities for different classes in the open-world setting, a reasonable assumption in many cases. However, in real-world applications, class distributions can be imbalanced, with some classes having an excessive number of instances, while others may have too few. Consequently, for broader practical applicability, TOC requires further optimization and deeper exploration to account for such variabilities.

\noindent
\textbf{FCID:} is specifically designed for tri-modal representations (audio, video, and text). However, in scenarios involving only two modalities, only the Fine or Coarse components of FCID’s disentanglement and alignment processes are applicable.

\section*{Acknowledgments}
We thank MindSpore (\url{http://mindspore.cn}) for the partial support of this work, which is a new deep learning computing framework.

% Bibliography entries for the entire Anthology, followed by custom entries
%\bibliography{anthology,custom}
% Custom bibliography entries only
\bibliography{acl_latex}

\appendix

\section{Implementation Details}
\label{sec:appendix_imple}
% \noindent
% \textbf{Downstream Details: }
% 所有的下游任务都是为了评估多模态统一表征对齐的程度，

\noindent
\textbf{Compared Works: }
 The models we compare include the most outstanding recent developments in multimodal unified discrete representations and models that excel in multimodal domain generalization: CODIS~\citep{duan2022multi}, TURN~\citep{zhao2022towards}, CMCM~\citep{liu2021cross}, SimMMDG~\citep{dong2024simmmdg}, and DCID~\citep{xia2024achieving}. These methods are implemented on our tasks, and their performance is evaluated on multi downstream tasks. 

\noindent
 \textbf{Evaluation Metrics: }
 For the AVE~\citep{avel}, VGGSound-AVEL~\citep{zhou2022contrastive, zhou2021positive}, and UCF101~\citep{soomro2012ucf101} datasets, precision is used as the metric. The F1-score is utilized for assessing the AVVP~\citep{tian2020unified} and AVE$\rightarrow$AVVP generalization task, and recall is utilized for zero-shot retrieval~\citep{chen2011collecting, drossos2020clotho}. Mean Square Error (MSE) is employed to evaluate the reconstruction quality of TOC on the CelebA-HQ 30K dataset~\citep{karras2017progressive}. Additionally, Fréchet Inception Distance (FID)~\citep{heusel2017gans} is used to assess the model's capability in cross-modal generalization.
\noindent
 \textbf{Hyperparameter Settings: }The $\beta$ of $L_{commit}$ is set to 0.25, and in the TOC formulation, the parameter $\lambda$ is set to 0.3, and in $L_{\text{nce}}$, the parameter $\tau$ is set to 1.0. All results presented in table~\ref{tab:cross-modal generalization}, \ref{tab:zero-shot retrieval}, \ref{tab:gen_result}, \ref{tab:Ablation of TOC}, \ref{tab:Ablation of FCID} were obtained with a codebook size set to 400 and an embedding dimension set to 256. The table~\ref{tab:vqvae_recon} involves VQVAE with a codebook size of 128 and an embedding dimension of 128. The ablation study on codebook size is discussed in Table~\ref{tab:Ablation of codebook size}. 

% \noindent
% \textbf{Computer resources: }Training the complete FCID model using a single Nvidia RTX 3090 GPU takes 10 hours, while TOC requires no additional training. Training the VQVAE model in this paper takes 1 hour on a single Nvidia RTX 3090. Fine-tuning related to cross-modal image generation requires 20 hours. All individual downstream experiments can be completed within 1 hour. The parameter count for the FCID Encoder (including the codebook) is 78M, while the DCID~\citep{xia2024achieving} Encoder (including the codebook) has 80M parameters. Compared to previous SOTA models, we achieve superior unified representation performance with a reduced parameter count.

\section{Experimental Supplement}
\label{sec:appendix_exper}

\noindent
\subsection{The Impact of TOC on Cosine Similarity of Codebook}
\label{sec:appendix_toc_cosine}
\begin{figure}[h]
  \centering
   \includegraphics[width=1.0\linewidth]{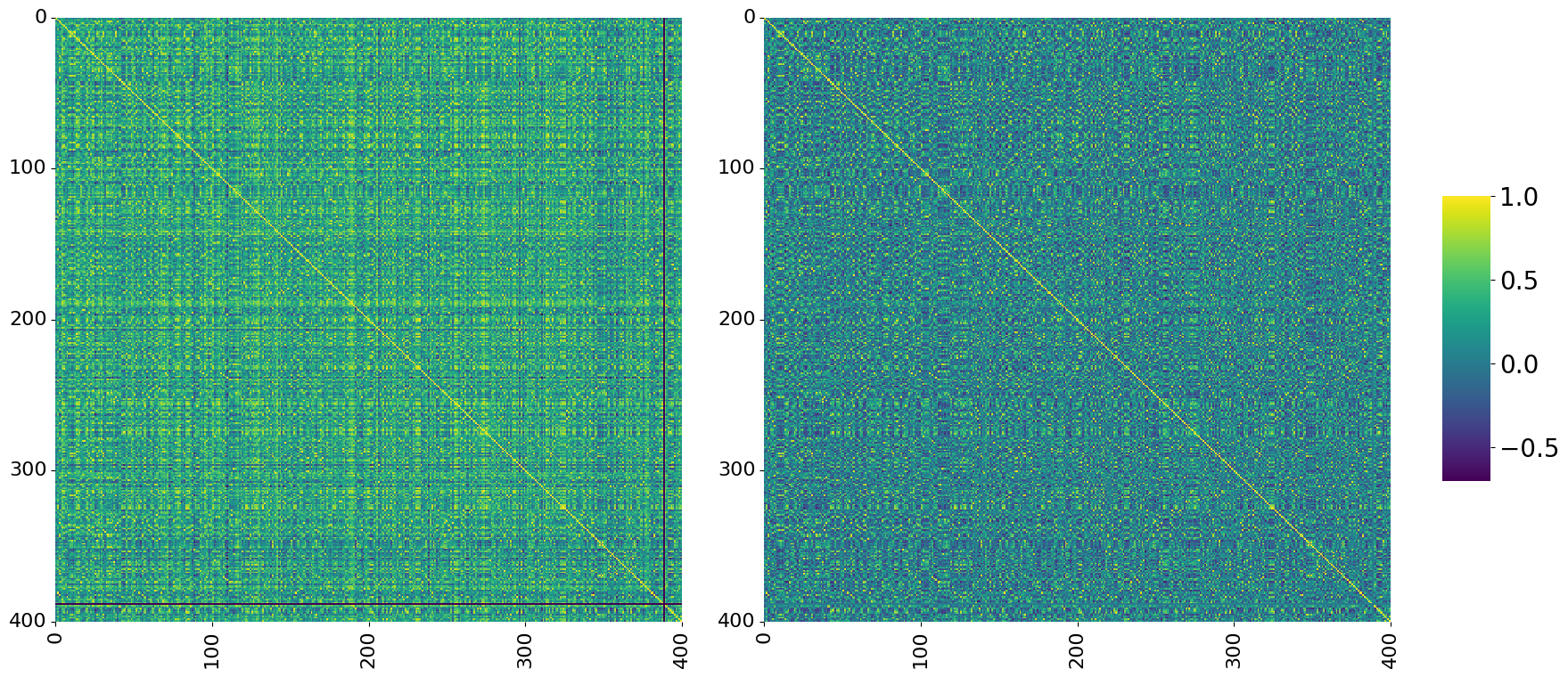}
   \caption{Left: Cosine similarity in the original codebook. Right: Cosine similarity after TOC.
   \label{fig:cosine}
   }
   \vspace{-3mm}
\end{figure}

We conducted an evaluation of TOC on the open-source pre-trained model of DCID~\citep{xia2024achieving}. As depicted in Figure~\ref{fig:cosine}, the distinctiveness of the codes with the features obtained after TOC computation is notably enhanced.

\noindent
\subsection{Computational efficiency}
\label{sec:appendix_computational_efficiency}

Comparing computational efficiency and resource requirements provides a more comprehensive view of the trade-off between performance improvement and computational cost. As shown in Table~\ref{tab:computational_efficiency}, we present a comparison of the computational efficiency of CMCM~\citep{liu2021cross}, DCID~\citep{xia2024achieving}, and our method:

\begin{table}[h!]
\centering
\resizebox{0.48\textwidth}{!}{
\begin{tabular}{cccc}
\toprule
\textbf{Model} & \textbf{GPU Memory Usage} & \textbf{Time per Epoch} & \textbf{Total Epochs} \\ \midrule

DCID & 100\% & 100\% & 5 \\ 
CMCM & 81\% & 76\% & 8 \\
FCID & 96\% & 99\% & 5 \\\bottomrule
\end{tabular}}
\caption{Model Comparison: Relative GPU Memory Usage and Time per Epoch (Compared to DCID), and Total Training Epochs}
\label{tab:computational_efficiency}
\end{table}

Under the same pretraining conditions with a batch size of 80 and using an RTX 3090 GPU, the results are shown in Table~\ref{tab:computational_efficiency}. In CMCM, contrastive learning is applied to both fine-grained and high-level representation for each modality. DCID performs alignment and disentanglement of all modalities at the fine-grained level (Alignment of pairwise combinations of the three modalities, along with their respective disentanglement). Our proposed FCID first applies disentangling and contrastive learning to fine-grained features for audio and video, then applies coarse-grained disentangling and contrastive learning to the compressed audio-visual and text features. Specifically, there are 2 alignment operations and 4 disentangling operations (fine-grained alignment of audio and video features, coarse-grained alignment of audiovisual and text features, and disentangling of these 4 feature types).

For GPU Memory Usage and Time per Epoch, CMCM has the lowest consumption in both cases. FCID uses less GPU memory than DCID, while Time per Epoch is nearly the same, mainly due to the projection compressing the temporal dimension before performing coarse-grained alignment and disentanglement, reducing computational data. DCID and FCID benefit from faster convergence due to disentangling and only require 5 epochs to achieve optimal performance, while CMCM needs warm-start strategy and takes 8 epochs to converge, resulting in the longest total training time.

During downstream inference, no disentangling or alignment is required. Only the corresponding encoders and quantization are used. Compared to DCID, our FCID adds only simple projection layers and coarse-grained general and specific encoders constructed with MLP for audiovisual data, making the added inference time almost negligible. The text modality, on the other hand, requires less time than DCID due to coarse-grained compression. The inference speed difference between CMCM, DCID, and FCID is minimal.

Another module we propose, TOC, only requires a single computation of less than 10 seconds after obtaining the discrete representations, and it only requires CPU computation. By reducing the feature dimensions, it can also accelerate training and inference speeds in downstream tasks.

\noindent
\subsection{Loss Changes During Pre-training}
\label{sec:appendix_loss}
\begin{figure}[ht]
  \centering
   \includegraphics[width=1.0\linewidth]{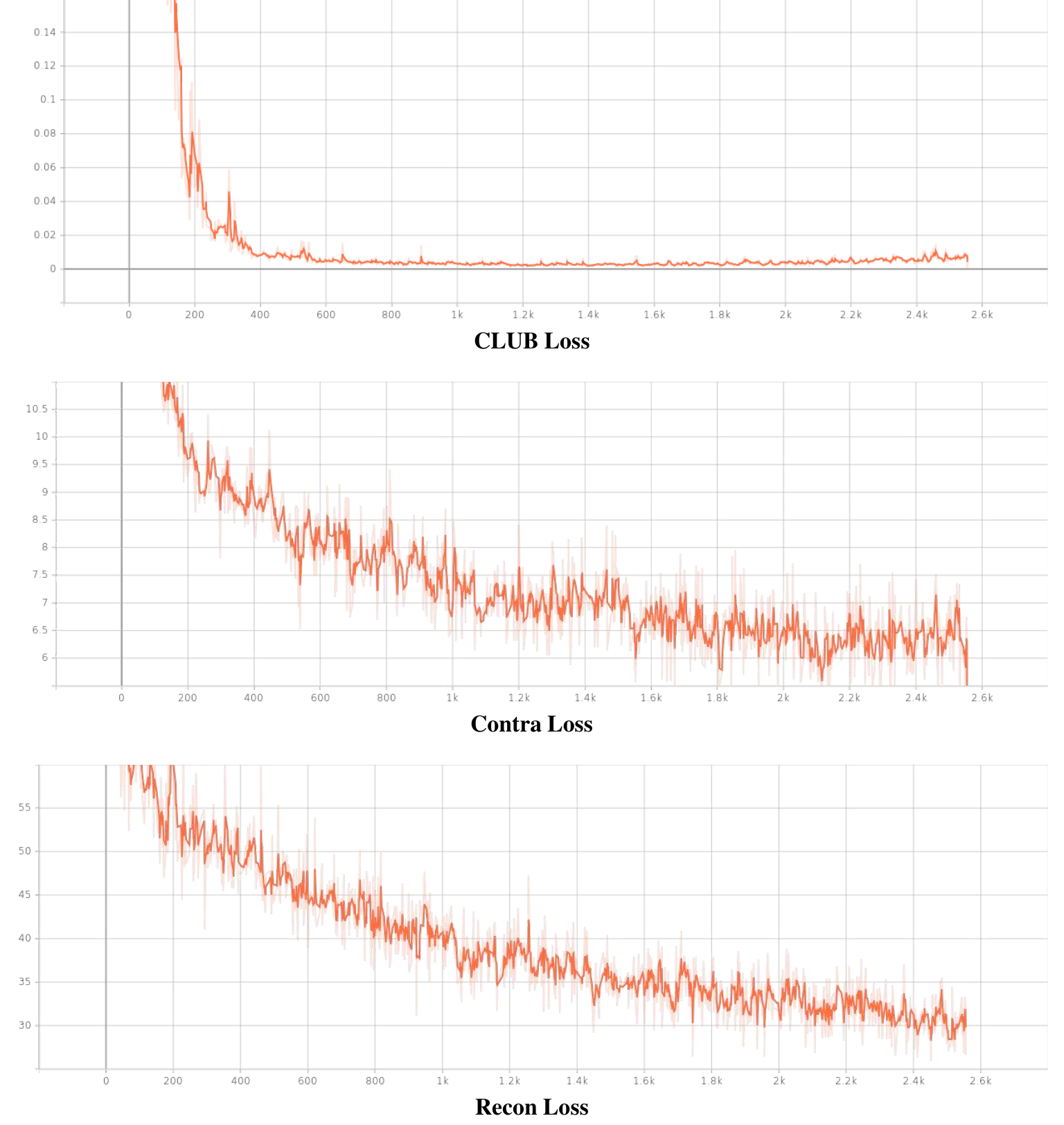}
   \caption{Changes in Losses During Pre-training
   \label{fig:loss_change}
   }
   \vspace{-3mm}
\end{figure}

As shown in Equation~\ref{eq19}, our model involves a total of four losses. Among them, $L_{commit}$ is a loss commonly used in VQ-related models. Its purpose is to ensure that the model's output maintains continuity in the latent space when mapped to the discrete space. This is not the main contribution of this paper. Here, we focus on the changes in the other three losses. As illustrated in the Figure~\ref{fig:loss_change}, the overall trend shows a clear downward slope. Among them, $L_{club}$ converges first, and when it reaches zero, it indicates successful decoupling of the modalities. The decrease in $L_{contra}$ reflects an increase in the alignment of multimodal information, while the decrease in $L_{recon}$ signifies improved preservation of complete semantics after the information is decoupled and re-merged.

% 如公式19所示，我们的模型一共涉及4个loss，其中$L_{commit}$是常用于VQ相关模型的loss，它的作用让模型的输出在映射到离散空间时，尽量保持其在潜在空间中的连续性，不是本文的主要贡献，我们这里主要列出其余3个loss的变化，如图所示，可以发现整体呈现明显下降趋势，其中$L_{club}$最先收敛，它收敛到0后说明各个模态信息解耦成功，$L_{contra}$下降代表多模态对齐程度上升，$L_{recon}$的下降代表信息解耦后再合并保留完整语义的程度提高。

\begin{table}[h]
% \vspace{-3mm}

\centering
\resizebox{0.48\textwidth}{!}{%
\begin{tabular}{ccccccccccc}
\toprule
\centering
\multirow{1}{*}{{Codebook Size}} 
                                 & 
                                 \multicolumn{2}{c}{\begin{tabular}[c]{@{}c@{}}AVE\\ V$\rightarrow$A   A$\rightarrow$V\end{tabular}} & 
                                 \multicolumn{2}{c}{\begin{tabular}[c]{@{}c@{}}AVVP\\ V$\rightarrow$A   A$\rightarrow$V\end{tabular}} & 
                                 \multicolumn{2}{c}{\begin{tabular}[c]{@{}c@{}}AVE$\rightarrow$AVVP\\ V$\rightarrow$A   A$\rightarrow$V\end{tabular}} & 
                                 \multicolumn{2}{c}{\begin{tabular}[c]{@{}c@{}}UCF(v)$\leftrightarrow$VGG(a)\\ V$\rightarrow$A   A$\rightarrow$V\end{tabular}}&
                                 \multicolumn{1}{c}{\begin{tabular}[c]{@{}c@{}}Avg.\\ \end{tabular}}\\
                                 \hline
256  & 52.9 & 52.3 & 38.8 & 43.2 & 53.7 & 53.9 & \textbf{70.8} & 56.4 & 52.75 \\
300  & 52.8 & 54.1 & 42.1 & 44.1 & 54.1 & \textbf{58.5} & 69.6 & 60.4 & 54.46 \\
400  & \textbf{55.2} & \textbf{54.9} & \textbf{42.4} & \textbf{44.5} & \textbf{55.3} & 57.4 & 69.4 & \textbf{61.6} & \textbf{55.09} \\
512  & 54.4 & 52.4 & 40.0 & 42.6 & 54.1 & 56.9 & 70.3 & 59.3 & 53.75 \\
800  & 52.2 & 54.6 & 41.6 & 43.9 & 53.1 & 56.7 & 69.6 & 59.7 & 53.93 \\
1024 & 52.8 & 54.5 & 40.4 & 41.6 & \textbf{55.3} & 55.9 & 65.8 & 58.6 & 53.11 \\
\bottomrule
\end{tabular}%
}
\caption{Ablation Studies on the Impact of Codebook Size}
\label{tab:Ablation of codebook size}
\end{table}

\begin{table*}[h]
\centering

\resizebox{0.95\textwidth}{!}{%
\begin{tabular}{ccccccccccccc}
\toprule
\centering
\multirow{2}{*}{Method} & \multicolumn{3}{c}{MSCOCO(V$\leftrightarrow$T)} & & \multicolumn{3}{c}{Clotho(A$\leftrightarrow$T)} & &\multicolumn{3}{c}{FlickrSound(V$\leftrightarrow$A)}  & \multirow{2}{*}{Avg.}
% \cline{2-4} \cline{6-8} \cline{10-12}
\\
% \cline{2-7} \cline{9-14}
 & R@1 & R@5 & R@10 & & R@1 & R@5 & R@10 & & R@1 & R@5 & R@10 & \\
\toprule
CMCM~\citep{liu2021cross} & 0.50 & 4.20 & 7.20 & & 1.62 & 8.04 & 14.87 & & 2.20 & 9.80 & 15.60 & 7.11  \\
DCID~\citep{xia2024achieving} & 0.80 & \textbf{5.00} & 8.30 & & 2.06 & 9.00 & 16.70 & & \textbf{3.10} & 11.10 & 17.20 & 8.14  \\
FCID & \textbf{1.30} & 4.90 & \textbf{9.60} & & \textbf{2.87} & \textbf{10.73} & \textbf{18.19} & & \textbf{3.10} & \textbf{11.80} & \textbf{17.50} & \textbf{8.89}  \\
\midrule
CMCM+TOC & 0.70 & 4.50 & 7.70 & & 1.93 & 8.43 & 15.33 & & 2.40 & 10.60 & 16.10 & 7.52\color{green}{(+0.41)}  \\
DCID+TOC & 1.10 & \textbf{5.30} & 8.80 & & 2.59 & 9.00 & 17.08 & & 3.60 & 11.80 & 17.80 & 8.56\color{green}{(+0.42)}  \\
FCID+TOC & \textbf{1.50} & 5.10 & \textbf{10.40} & & \textbf{3.16} & \textbf{11.15} & \textbf{19.04} & & \textbf{3.80} & \textbf{12.20} & \textbf{18.40} & \textbf{9.42}\color{green}{(+0.53)}\\
\bottomrule
\end{tabular}
}
\caption{Details of comparison with SOTA methods on three cross-modal zero-shot retrieval tasks, all results are calculated as the mean across two directions.}
\label{tab:Details of zero-shot retrieval}
\end{table*}

\begin{table*}[h]
% \vspace{-3mm}

\centering
\resizebox{0.95\textwidth}{!}{%
\begin{tabular}{cccccccccccccc}
\toprule
\centering
\multirow{1}{*}{{$A_{CLUB}$}} 
& \multirow{1}{*}{{$V_{CLUB}$}} 
& \multirow{1}{*}{{$AV_{CLUB}$}} 
& \multirow{1}{*}{{$TE_{CLUB}$}} 
                                 & 
                                 \multicolumn{2}{c}{\begin{tabular}[c]{@{}c@{}}AVE\\ V$\rightarrow$A   A$\rightarrow$V\end{tabular}} & 
                                 \multicolumn{2}{c}{\begin{tabular}[c]{@{}c@{}}AVVP\\ V$\rightarrow$A   A$\rightarrow$V\end{tabular}} & 
                                 \multicolumn{2}{c}{\begin{tabular}[c]{@{}c@{}}AVE$\rightarrow$AVVP\\ V$\rightarrow$A   A$\rightarrow$V\end{tabular}} & 
                                 \multicolumn{2}{c}{\begin{tabular}[c]{@{}c@{}}UCF(v)$\leftrightarrow$VGG(a)\\ V$\rightarrow$A   A$\rightarrow$V\end{tabular}}&
                                 \multicolumn{1}{c}{\begin{tabular}[c]{@{}c@{}}Avg.\\ \end{tabular}}\\
                                 \hline
- & - &- & - &51.3 & 51.6 & 39.5  & 40.7  & 50.6 & 51.1 & 63.3    & 57.6 & 50.71\\ 
\checkmark & - &- & - &52.4 & 53.5 & 40.9  & 42.4  & 53.1 & 54.2 & 66.0    & 59.8 & 52.79\\ 
- & \checkmark &- & - &53.1 & 53.4 & 41.7  & 43.2  & 53.9 & 54.7 & 67.1    & 60.1 & 53.40\\ 
- & - &\checkmark & - &52.2 & 51.9 & 40.2  & 41.7  & 52.4 & 52.5 & 64.2    & 59.1 & 51.78\\ 
- & - &- & \checkmark &51.7 & 51.5 & 40.6  & 41.8  & 52.5 & 52.9 & 63.5    & 58.2 & 51.59\\ 
\checkmark & \checkmark &- & -  &54.2 & 54.0 & 41.4  & 43.9  & \textbf{55.9} & 56.1 & 67.9    & 61.3 & 54.34\\ 
- & - &\checkmark & \checkmark  &52.9 & 52.6 & 40.8  & 42.1 & 52.5 & 53.9 & 65.7 & 59.2   & 52.46\\ 
\checkmark & \checkmark &\checkmark & \checkmark  & \textbf{55.2} & \textbf{54.9} & \textbf{42.4} & \textbf{44.5} & 55.3 & \textbf{57.4} & \textbf{69.4} & \textbf{61.6} & \textbf{55.09}\\ 
\bottomrule
\end{tabular}%
}
\caption{Details of ablation studies on the impact of FCID}
\label{tab:Details of Ablation of FCID}
\end{table*}

\noindent
\subsection{Codebook size}
\label{sec:appendix_codebooksize}

Table~\ref{tab:Ablation of codebook size} presents the performance of the FCID model across various codebook sizes. It is observed that the model achieves the best average results when the codebook size is set to 400. Conversely, using either a excessively large or small codebook size may lead to insufficient semantic learning or inadequate semantic expression, resulting in decreased model performance.

\subsection{Reconstruction}
\label{sec:appendix_recon}

\begin{figure}[h]
  \centering
  \setlength{\abovecaptionskip}{0.5mm}
   \includegraphics[width=1.0\linewidth]{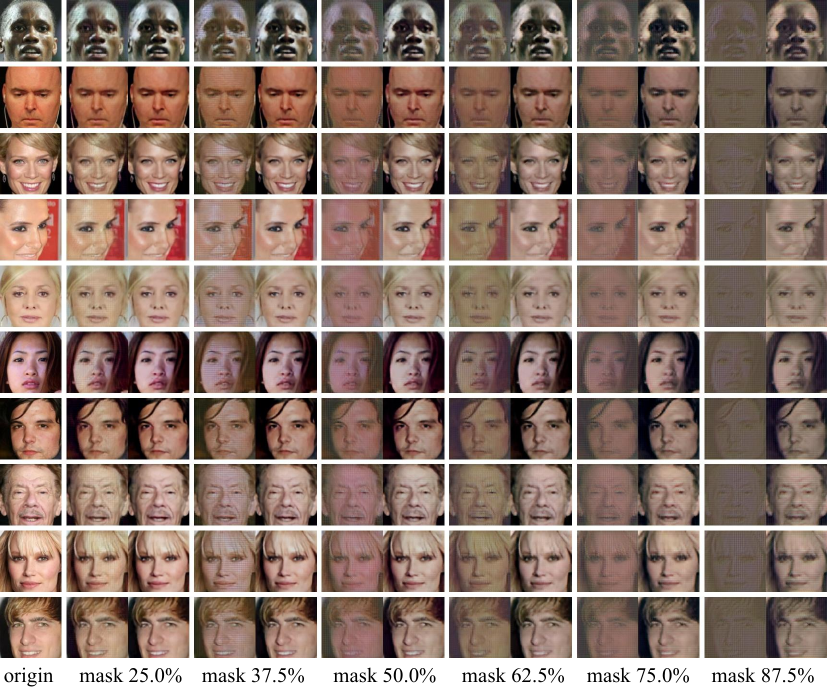}
   \caption{More results of reconstructions using random and TOC masking.
   \label{fig:appendix_vqvae_recon_sample}
   }
\end{figure}

% As shown in Figure~\ref{fig:appendix_vqvae_recon_sample}, 除了'origin'列，其余列左侧图像为随机掩码重构，右侧为TOC掩码保留分数最高的dimensions重构。可以发现在25\%到87.5\%的掩码重构中，TOC的表现明显由于随机掩码，并且随着掩码比例的增加差距越发明显。
As shown in Figure~\ref{fig:appendix_vqvae_recon_sample}, for all columns except the 'origin' column, the images on the left represent reconstructions with random masks, while the images on the right illustrate reconstructions using the dimensions with the highest TOC retention scores. It is evident that TOC significantly outperforms random masking in reconstructions with mask ratios ranging from 25.0\% to 87.5\%, with the performance gap becoming increasingly pronounced as the mask ratio increases.

\begin{figure}[b]
  \centering
   \includegraphics[width=1.0\linewidth]{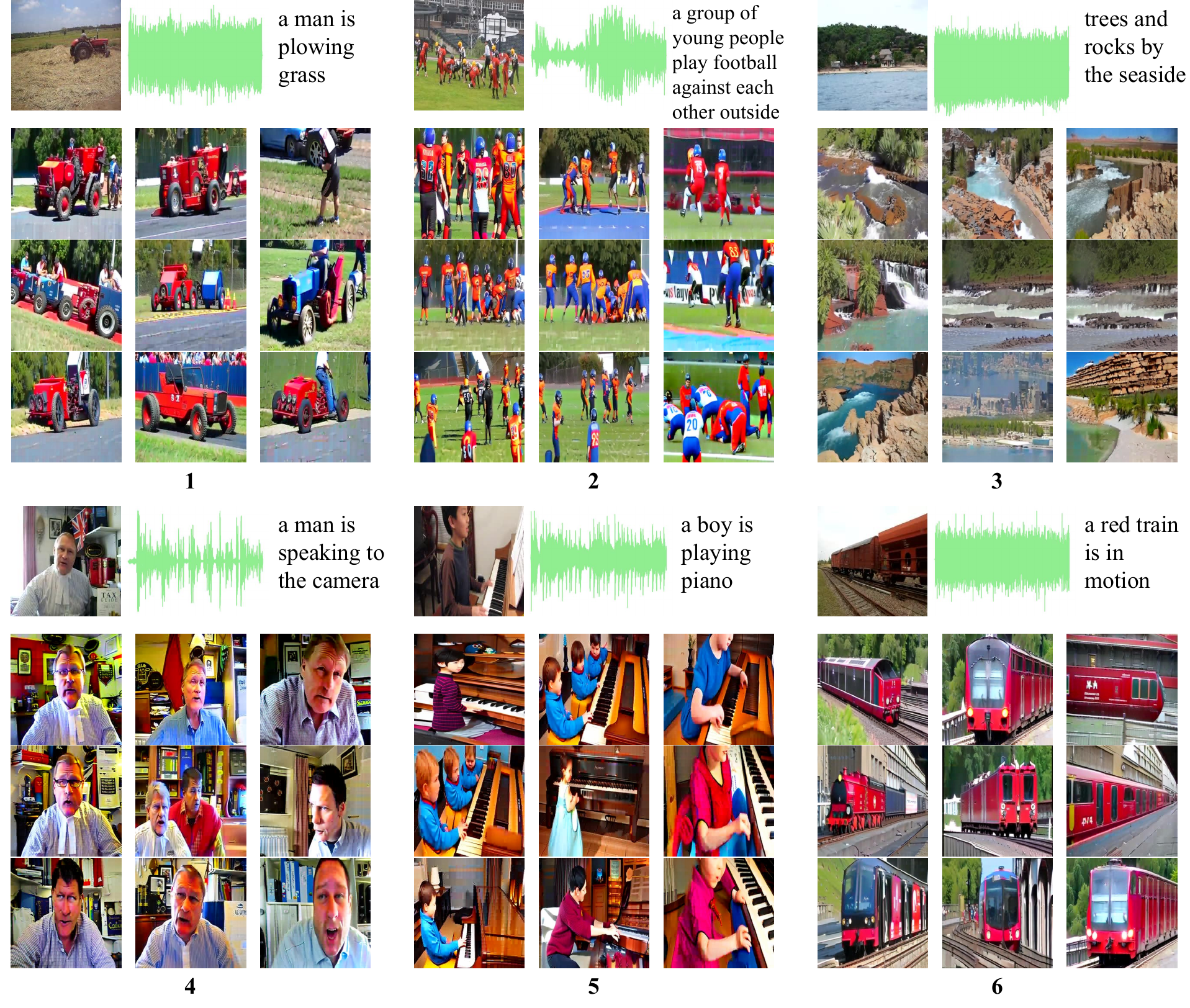}
   \caption{More results of cross-modal generation.
   \label{fig:appendix_crossmodal_gen}
   }
\end{figure}

\subsection{Generation}
\label{sec:appendix_gen}

As shown in Figure~\ref{fig:appendix_crossmodal_gen}, thanks to multimodal unified representations, the results of cross-modal image generation from audio and text closely resemble actual images. As evident in samples 2 and 6, despite the audio not mentioning specific details such as the color of clothing and trains, these elements are still accurately generated, which can be attributed to the discrete unified representation serving as a central semantic hub for multiple modalities. In contrast, the results from Text-to-Image (T$\rightarrow$I) are noticeably inferior to those from Image-to-Image (I$\rightarrow$I) and Audio-to-Image (A$\rightarrow$I). This difference is exemplified in the first image generated from sample 1's text, where the action of a car mowing grass is mistakenly transformed into a man mowing grass. This discrepancy arises because the semantic connections between images and audio are stronger than those generated through model-based text, which merely mentioned 'man' and 'plowing grass' without specifying the tool used for plowing.

% 得益于多模态统一表征，音频和文本跨模态生成图像的结果能做到接近图像，如样本2和样本6，音频中没有提到的服装颜色和火车颜色仍然能被生成出来，这一定程度受益于离散的统一表征作为多个模态的中心语义。而T$\rightarrow$I的结果明显不如I$\rightarrow$I和A$\rightarrow$I的原因在样本1样本1的文本生成的第一张图上也能体现，它将原本的车子除草变成了人除草。这是因为图像和音频的语义联系远比通过模型生成的文本的语义联系更高，可以看到文本中只提到了'man'和'plowing grass'，而没有提到除草用的工具。

\end{document}